\documentclass{article} % For LaTeX2e
\usepackage{iclr2027_conference,times}

\usepackage{amsmath,amsfonts,bm}

\def\eqref#1{equation~\ref{#1}}
\def\1{\bm{1}}

\DeclareMathAlphabet{\mathsfit}{\encodingdefault}{\sfdefault}{m}{sl}
\SetMathAlphabet{\mathsfit}{bold}{\encodingdefault}{\sfdefault}{bx}{n}

\usepackage{hyperref}
\usepackage{url}

\usepackage{graphicx}
\usepackage{subcaption}
\usepackage{enumitem}
\usepackage{xcolor}
\usepackage{amsmath,amssymb,amsthm}
\usepackage{pifont}
\usepackage{booktabs}
\usepackage{multirow}
\usepackage[table]{xcolor}
\usepackage{colortbl}
\usepackage{array}
\usepackage{tabularx}
\usepackage{makecell}
\usepackage{caption}
\usepackage{enumitem}
\usepackage{algorithm}
\usepackage{algpseudocode}

\newcommand{\mydashline}{%
\noalign{%
\vskip-0.15pt
\hbox to \linewidth{%
\leaders\hbox{\rule{2.5pt}{0.3pt}\hskip2pt}\hfill
}%
\vskip-0.15pt
}%
}

\definecolor{BestYellow}{RGB}{246, 224, 164}
\definecolor{SecondYellow}{RGB}{250, 241, 211}
\definecolor{HeaderOrange}{RGB}{248, 226, 210}

\definecolor{TeacherBlue}{RGB}{205, 225, 242}
\newcommand{\best}[1]{\cellcolor{BestYellow}{\bfseries\boldmath #1}}
\newcommand{\second}[1]{\cellcolor{SecondYellow}#1}
\title{Dual-Vocabulary Language Model for Cross-tokenizer Distillation}

\author{
 \textbf{Kedi Chen\textsuperscript{1,3}},
 \textbf{Chen Lin\textsuperscript{1}},
 \textbf{Yutao Sun\textsuperscript{2,3}},
 \textbf{Wei Zhang\textsuperscript{1,3}\thanks{Corresponding author.}}
\\
 \textsuperscript{1}East China Normal University,
 \textsuperscript{2}Tsinghua University,
 \textsuperscript{3}Shanghai Innovation Institute
 \\
 \textbf{Email Contact:} kdchen2@stu.ecnu.edu.cn, zhangwei.thu2011@gmail.com
\\
\\
}

\iclrfinalcopy % TODO

\begin{document}

\maketitle
\fancyhead[L]{} % TODO

\begin{abstract}
On-policy distillation (OPD) bridges teacher supervision and student behavior, but different teacher-student tokenizers introduce misalignment in both input tokenization (\#1) and output logits (\#2).
Existing approaches address the former by matching same-text spans or converting tokens to bytes, often losing fine-grained token information or disrupting the native-token paradigm, while for the latter, strategies such as ranking, padding, or key-token selection retain only shared logit dimensions, resulting in much distribution loss.
In this paper, we propose Dual-Vocabulary Language Model (DVLM), which replaces the teacher’s LM head with a new student-vocabulary projection head and obtains full-dimensional student logits (for \#2). 
To support student tokens (for \#1), it takes a Parallel-Tokenized Sequence (PTS) as input, which concatenates the original teacher-tokenized sequence and a re-tokenized sequence formed by independently converting each student token into a teacher-token group.
To avoid inference inconsistency with the original teacher tokens, the Hybrid-Prefix Attention (HPA) further restricts re-tokenized groups to their corresponding teacher prefix and uses its last state as the aggregation of the original student-token representation for projection into the student vocabulary space.
Similarly, via the combined use of PTS and HPA, the DVLM teacher can provide distribution-aligned supervision with the student’s input-tokenization and output-logit during OPD.
Experimental results demonstrate that our DVLM teacher has a similar converged loss as the original teacher model and enables student models to improve performance across six reasoning tasks.
% Our code and data: \textcolor{blue}{https://anonymous.4open.science/r/DVLM-C0B7/}.
\end{abstract}

\section{Introduction}
\label{sec: intro}
On-policy distillation (OPD) \citep{yang2025qwen3technicalreport,song2026surveyonpolicydistillationlarge} receives increasing attention in recent years for leveraging on-policy student-generated responses to mitigate the distribution mismatch between teacher supervision and student behavior.
In practice, however, teacher and student models are often equipped with different tokenizers \citep{jia2025principlesapplicationscomprehensivesurvey}, making cross-tokenizer distillation an important setting for OPD in large language models (LLMs) \citep{Zhao_2026}.

\begin{figure*}[t]
    \centering

    \begin{subfigure}[t]{0.63\textwidth}
        \centering
        \includegraphics[width=\linewidth]{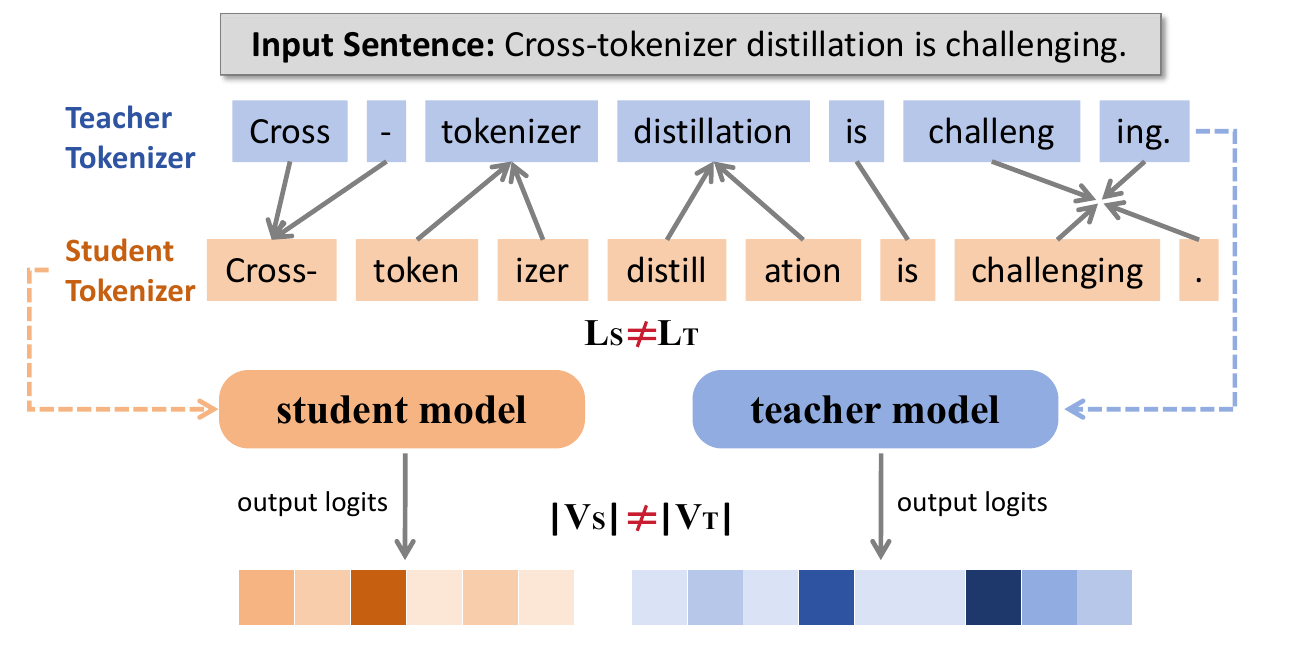}
        \caption{Misalignments on input tokenization and output logits.}
        \label{fig:sub_a}
    \end{subfigure}
    \hfill
    \begin{subfigure}[t]{0.33\textwidth}
        \centering
        \includegraphics[width=\linewidth]{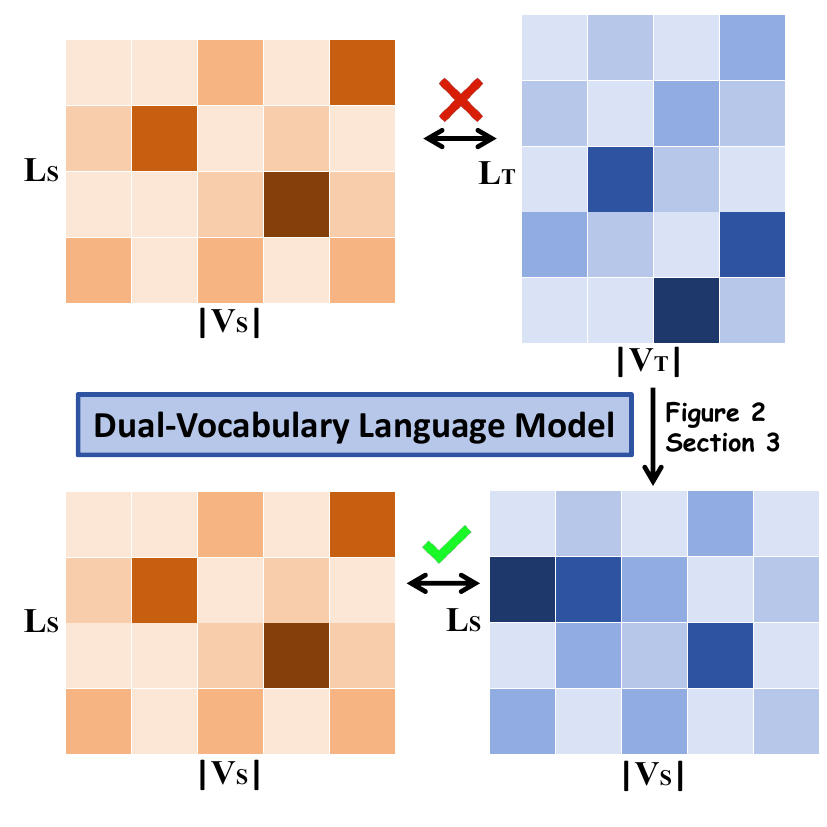}
        \caption{Our method helps alignment.}
        \label{fig:sub_b}
    \end{subfigure}

    \caption{(a) The key obstacle in cross-tokenizer distillation is that the teacher and student tokenizers segment the same text differently ($L_S \neq L_T$), while their distinct vocabularies yield logits of different dimensionalities ($|V_S| \neq |V_T|$). (b) Our method DVLM can make the teacher match the student’s tokenization schema and produce logits with the same dimensionality through the PTS input format and HPA mechanism, thereby achieving distributional alignment. }
    \label{fig:1}
\end{figure*}

The core challenges in cross-tokenizer distillation lie in \textit{aligning the teacher-student output distributions across input tokenization \text{(\#1)} and output logits \text{(\#2)}}.
The former arises because different tokenizers produce different token boundaries for the same input text \citep{DBLP:conf/naacl/AliFTRLLKEDBJWJAJSOWSKF24}, while the latter stems from LLMs having different output logit spaces over different vocabularies \citep{DBLP:conf/emnlp/LiuWQKKS024}.
Figure~\ref{fig:sub_a} illustrates the misalignment in these two aspects.

Existing studies introduce several approaches for these challenges.
For (\#1) input-tokenization alignment, prior methods mostly match teacher and student tokens that correspond to the same text span \citep{minixhofer2025universalcrosstokenizerdistillationapproximate,sreenivas2026xtokenprojectionguidedcrosstokenizerknowledge}, thereby recasting token-level distributions as span-level distributions, or directly convert the common token format into the byte format to enable tokenizer-independent distillation \citep{phan2026crosstokenizerlikelihoodscoringalgorithms,wang2026crosstokenizeronpolicydistillationbyteprefix}.
These techniques either conduct token-to-span distribution aggregation at the cost of fine-grained token information loss, or heavily rely on byte-level splitting of native tokens, which substantially deviates from LLMs’ natural pretraining paradigm.
% Fundamentally, they deviate from the original training and inference paradigm of the LLM backbones.
For (\#2) output-logit alignment, current research generally performs operations such as ranking, padding, and key-token exploration based on token probabilities to identify a subset of shared logit dimensions across different vocabularies \citep{boizard2025crosstokenizerdistillationuniversallogit,nguyen2026ctpdcrosstokenizerpreference}. 
Given that modern LLM vocabularies often exceed $100$K tokens and differ largely from one another, these strategies lead to much distribution loss in the non-shared logit dimensions.

Since prior approaches cannot fundamentally resolve tokenization and logit-space issues, we propose a two-step method for cross-tokenizer OPD.
Stage I: We replace the teacher's original LM head \citep{Shao_2024} with a new projection head that can map to the student vocabulary and obtain full-dimensional student logits (for \#2).
We refer to this architecture as \textbf{D}ual-\textbf{V}ocabulary \textbf{L}anguage \textbf{M}odel (\textbf{DVLM}).
To make the DVLM teacher compatible with the student tokenization schema, we allow it to `accept' student tokens (for \#1): 
The new teacher DVLM takes a \textbf{Parallel-Tokenized Sequence} (PTS) as input, which concatenates two tokenized sequences of the same text: the original teacher-tokenized sequence and a re-tokenized sequence formed by independently converting each student token into a teacher-token group.
To avoid the inference inconsistency between the re-tokenized tokens and the original teacher tokens, we further introduce a \textbf{Hybrid-Prefix Attention} (HPA) mechanism, under which each re-tokenized token can attend only to the corresponding prefix of the teacher-tokenized sequence. 
The representation at the last position of each re-tokenized group can be viewed as an aggregation of the original student token, and is then projected into the student vocabulary space.
During training, the teacher backbone is frozen, and only the new lightweight projection head is optimized.
Stage II: Once the DVLM teacher is trained, it can be directly used for OPD. 
For each student-generated response, we also construct the PTS with HPA to obtain the corresponding teacher logits. 
Similarly, the logits at the last position of each re-tokenized group are used to establish a one-to-one correspondence with the student token positions.
In this way, the teacher DVLM is consistent with the student model in both input tokenization and the output logits, thereby achieving distribution-dimensional alignment (Figure~\ref{fig:sub_b}).
The student model is then optimized under the aligned teacher by minimizing the forward KL divergence.

To assess our method, we select two teacher-student pairs; thus, two DVLM teachers are obtained.
For Stage I, we show through loss convergence and results on four language-understanding benchmarks that our DVLM teacher reaches the performance level of the original teacher model. 
For Stage II, we evaluate the distilled student model on six math and code reasoning benchmarks, where it outperforms conventional post-training methods and existing cross-tokenizer distillation methods.

The main contributions of this work are summarized as follows:
\begin{itemize}[leftmargin=1.2em, itemsep=2pt, topsep=2pt]
\item We propose a Dual-Vocabulary Language Model for cross-tokenizer distillation, a teacher with a new LM head that can project the logits into the student vocabulary space (for \#2).
\item We introduce a Parallel-Tokenized Sequence and a Hybrid-Prefix Attention mechanism for one-to-one teacher–student token mapping in DVLM teacher training and OPD (for \#1).
\item Experiments demonstrate the DVLM teacher has a similar converged loss as the original teacher model and enables student models to improve performance across diverse reasoning tasks.
\end{itemize}

\section{Related Work}
\label{sec: related}
\paragraph{On-Policy Distillation.}
On-policy distillation (OPD) \citep{zhang2026formuladrivensurveyresearchagenda} is an emerging knowledge distillation mode in which the student model generates its own response trajectories and learns from teacher feedback on the logit states \citep{agarwal2024onpolicydistillationlanguagemodels,gu2026minillmonpolicydistillationlarge}.
Unlike conventional off-policy distillation that relies primarily on fixed teacher-generated or ground-truth text, OPD can alleviate the training-inference distribution mismatch and exposure bias of autoregressive language models \citep{ko2024distillmstreamlineddistillationlarge,lin2026renioreweightingnegativetrajectory}.
Our paper focuses on OPD in cross-tokenizer settings.

\paragraph{Cross-tokenizer Distillation.} 
Compared with common OPD, the key challenge in cross-tokenizer distillation is that the teacher and student exhibit misalignment in both their input tokenization and output logits \citep{song2026surveyonpolicydistillationlarge}.
(1) For input-tokenization alignment, 
some early studies use optimal transport or edit distances to measure pairwise distances between tokens \citep{le2025cot2aligncrosschainthoughtdistillation,DBLP:conf/aaai/VuongLTVL26}.
Later, one common approach \citep{DBLP:conf/icml/Shin0LG25,sun2026simctrecoveringlostsupervision} is to identify shared text spans across different tokenizers and aggregate token-level distributions into span-level distributions using joint probability.
Another line of work converts tokens into byte sequences. 
It uses a prefix trie to marginalize token probabilities into next-byte conditional distributions, thereby aligning the teacher and student in a shared byte space \citep{phan2026crosstokenizerlikelihoodscoringalgorithms,wang2026crosstokenizeronpolicydistillationbyteprefix}.
(2) For output-logit alignment, current strategies generally manipulate the output distributions through operations such as ranking \citep{cui2025multileveloptimaltransportuniversal,sreenivas2026xtokenprojectionguidedcrosstokenizerknowledge}, padding \citep{boizard2025crosstokenizerdistillationuniversallogit}, and key-token exploration \citep{DBLP:conf/icml/Shin0LG25,nguyen2026ctpdcrosstokenizerpreference} based on token probabilities, or divide the vocabularies into matched and unmatched tokens \citep{patiño2025_unlocking_on_policy_distillation_for_any_model_family}, and identify a subset of shared logit dimensions across different vocabularies.
As discussed above, both types of methods incur greater distribution loss and disrupt the native-token paradigm. 
Therefore, this paper constructs a PTS that enables the teacher to `accept' student tokens and trains the DVLM architecture to output student logits, aligning their distributions during distillation.

\section{Method}
\label{sec: method}

\begin{figure*}[t]
    \centering
    \scalebox{0.95}[0.9]{
        \includegraphics[width=\textwidth]{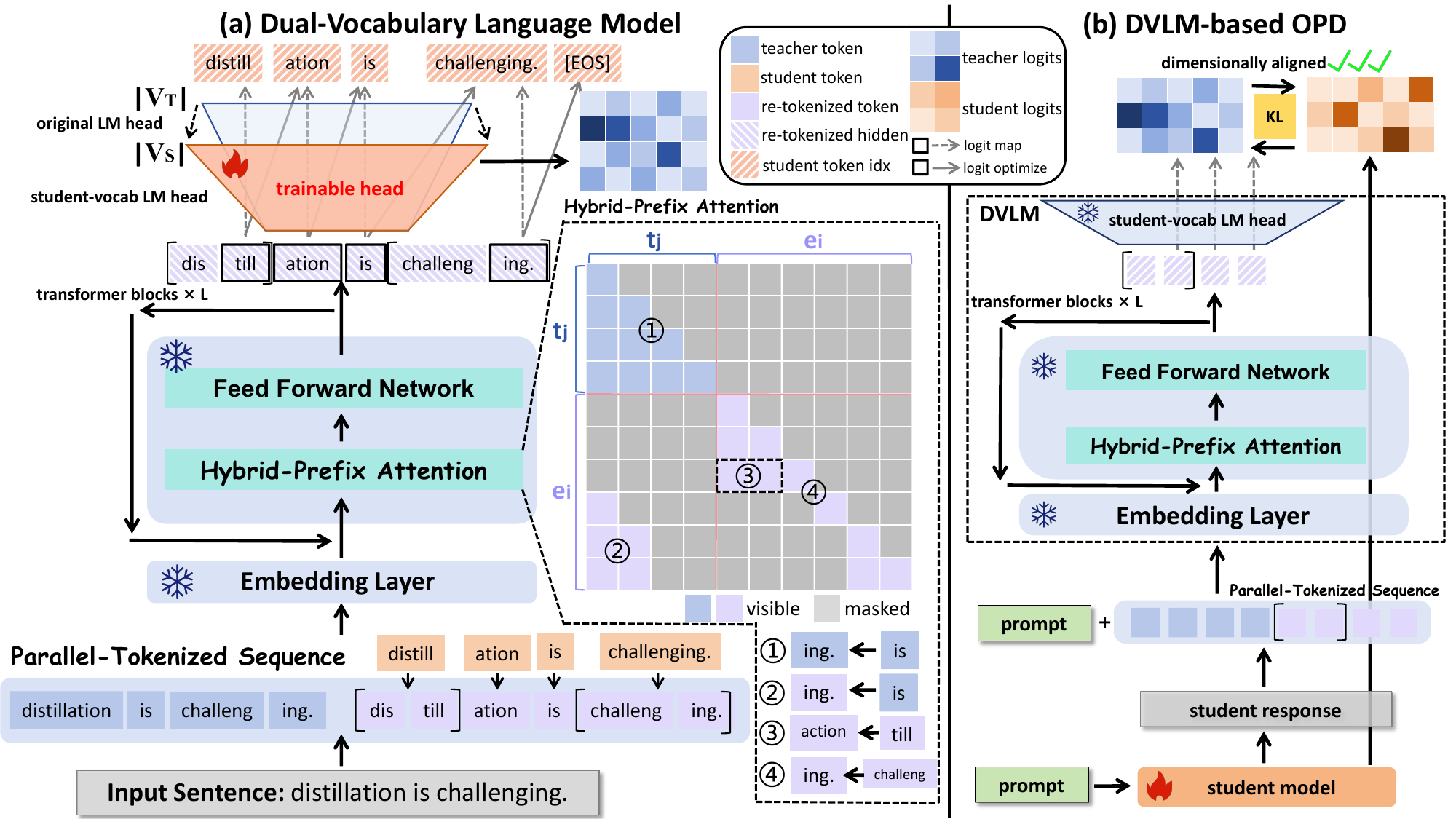}
    }
    \caption{(a) The Dual-Vocabulary Language Model takes the Parallel-Tokenized Sequence as input, with its token visibility controlled by Hybrid-Prefix Attention, while only the student-vocabulary LM head is trained. (b) In the same way, the trained DVLM teacher can extract output distributions with exactly the same dimensionality as those of the student model for OPD.}
    \label{fig:main}
\end{figure*}

The workflow of our method is in Figure~\ref{fig:main}, which consists of two steps: 
(a) training a new LM head on top of the teacher model backbone to map its representations into the student's vocabulary space; and (b) using the DVLM as the new teacher to guide the student through OPD.
We will use the example in Figure~\ref{fig:main} (in \underline{italic} form) to explain each step of our method in detail.

\subsection{Dual-Vocabulary Language Model}
\label{sec: DVLM}
\paragraph{Overview.} 
The goal of training the DVLM teacher is to enable the teacher backbone to produce logits over the student vocabulary. 
Thus, we design the Parallel-Tokenized Sequence (PTS) as the input to the DVLM architecture and further introduce the Hybrid-Prefix Attention (HPA) mechanism to control the visibility among all the tokens. 

Given an input text $x$, the teacher and student tokenizers, $\tau_T(\cdot)$ and $\tau_S(\cdot)$, produce two token sequences: with $L_T$ tokens and $L_S$ tokens (\textit{blue and orange tokens}), respectively.
\begin{equation}
\tau_T(x)
=
(t_1,\ldots,t_{L_T}),
\qquad
\tau_S(x)
=
(s_1,\ldots,s_{L_S})
\end{equation}

To address the mismatch between teacher and student tokenization schema in cross-tokenizer OPD, we enable the teacher model to `accept' the student tokens corresponding to the same input text.
Each student token $s_i$ is then independently decoded and re-tokenized to $m_i$ teacher tokens via $\tau_T(\cdot)$, forming a re-tokenized group $\mathbf{e}_i$:
\begin{equation}
\mathbf{e}_i
=
\tau_T\!\left(
\operatorname{Decode}(s_i)
\right)
=
(e_{i,1},\ldots,e_{i,m_i}),
\qquad
i=1,\ldots,L_S
\end{equation}
In the figure, when $i = 1$, the student token \textit{`distill'} is re-tokenized to two teacher tokens \textit{`dis'} and \textit{`till'} (\textit{in purple}).
These two tokens constitute a re-tokenized group $\mathbf{e}_1$.

In this way, all student tokens $s_i$ are transformed into teacher tokens that can be fed into the DVLM teacher. 
However, these re-tokenized groups do not follow the teacher’s native tokenization schema, which may introduce train–inference inconsistency in the teacher backbone \citep{agarwal2024onpolicydistillationlanguagemodels}. 
Thus, we introduce PTS together with HPA mechanism. 
The key idea is to preserve the teacher’s native tokenization as prefix context, maximizing the use of its original tokenization schema.
Specifically, we prepend the original teacher sequence $\tau_T(x)$ to the re-tokenized groups to construct the complete PTS, named $\mathbf{X}_{\mathrm{PTS}}$:
\begin{equation}
\label{eq: pts}
\mathbf{X}_{\mathrm{PTS}}
=
[(t_1,\ldots,t_{L_T});(\mathbf{e}_1,\ldots,\mathbf{e}_{L_S})]
\end{equation}
where $[;]$ denotes concatenation. 
We will next introduce the definition of Token Alignment Group and describe the HPA mechanism thoroughly.

\paragraph{Definition: Token Alignment Group.}
Given two token sequences $\tau_T(x)$ and $\tau_S(x)$ of the same text, we greedily aggregate consecutive tokens until their decoded text spans match. 
Let $\mathcal{G}_k^T$ and $\mathcal{G}_k^S$ denote the $k$-th teacher- and student-aligned groups, respectively, where they meet $\operatorname{Decode}(\mathcal{G}_k^T)=\operatorname{Decode}(\mathcal{G}_k^S)$. 
For example, when $k = 1$, $\mathcal{G}_1^T = \left[`distillation'\right]$ and $\mathcal{G}_1^S = \left[`distill', `ation'\right]$, the two share the same text span `distillation'.
Accordingly, for $t_j\in\mathcal{G}_k^T$ and $s_i\in\mathcal{G}_k^S$, their group indices satisfy $g_T(j)=g_S(i)=k$. 
For $t_1=`distillation'$ and $s_1=`distill'$, $g_T(1)=g_S(1)=1$. 
See Appendix~\ref{app: token alignment group} for the computation procedure and further details.

Next, we describe how HPA works.
Let $M(a,b)=1$ indicate that query token $a$ can attend to key token $b$ in the attention matrix $M$:
\begin{equation}
\label{eq:4}
M(a,b)=
\begin{cases}
1, & a=t_j,\ b=t_\ell,\ \ell\leq j,\\
1, & a=\mathbf{e}_{i,r},\ b=t_j,\ g_T(j)<g_S(i),\\
1, & a=\mathbf{e}_{i,r},\ b=\mathbf{e}_{u,v},\
     g_S(u)=g_S(i),\ u<i,\\
1, & a=\mathbf{e}_{i,r},\ b=\mathbf{e}_{i,v},\ v\leq r,\\
0, & \text{otherwise}.
\end{cases}
\end{equation}
Specifically, the original teacher sequence follows the normal causal attention (e.g., $t_j=`is',t_\ell=`distillation'$, \textit{area \ding{172}}). 
In contrast, each re-tokenized token in $e_{i,r}$ (e.g., $e_{i,r}=`ing.'$) attends to:
(i) all original teacher tokens $t_j$ in the alignment groups preceding (\textit{two teacher tokens: `distillation' and `is', not `challeng', because `challeng' and `ing.' are both in $\mathcal{G}_3^T$ or $\mathcal{G}_3^S$}, \textit{area \ding{173}}); 
(ii) all re-tokenized tokens of earlier student tokens within the same alignment group (\textit{no earlier student tokens in this example}, \textit{area \ding{174}}); 
and (iii) itself and its preceding tokens within $\mathbf{e}_i$ (\textit{two re-tokenized tokens: `ing.' itself and `challeng' are both in $\mathbf{e}_4$}, \textit{area \ding{175}}). 

After HPA, for each student token $s_i$, we extract the hidden state of its last re-tokenized token $e_{i,m_i}$, denoted by $\mathbf{h}^{e}_{i,m_i}$, as its teacher-side representation (\textit{the purple diagonally striped tokens outlined in black}).
Each original student token can be paired with one corresponding hidden state in the DVLM teacher, ensuring consistency with the student tokenization schema.
The new LM head of the DVLM teacher $\mathbf{W}_{D}$ ($\mathbb{R}^{d \times |V_S|}$) projects $\mathbf{h}^{e}_{i,m_i}$ ($\mathbb{R}^{d}$) onto the student vocabulary space and is optimized to predict the next student token $s_{i+1}$ (\textit{the orange diagonally striped tokens}).
\begin{equation}
\mathbf{W}_D^T \mathbf{h}^{e}_{i,m_i}
\in \mathbb{R}^{|V_S|},
\qquad i=1,\ldots,L_S.
\end{equation}

To retain the teacher's performance maximally, the new LM head $\mathbf{W}_D$ is initialized from the teacher LM head $\mathbf{W}_T$. 
The teacher-student shared tokens directly inherit their teacher weights, while unmatched student tokens are initialized by averaging the weights of their re-tokenized teacher tokens.
\begin{equation}
\mathbf{W}_D[:,i]
=
\begin{cases}
\mathbf{W}_T[:,j], & \text{if } s_i=t_j\in V_T, \\
\displaystyle
\frac{1}{m_i}\sum_{r=1}^{m_i} \mathbf{W}_T[:,\mathbf{e}_{i,r}],
& \text{otherwise},
\end{cases}
\qquad i=1,\ldots,|V_S|.
\end{equation}
During training, we freeze the teacher backbone and only train lightweight LM heads (training time explanation in
Appendix~\ref{app: more training}) for different student vocabularies.
The standard cross-entropy loss for LLM training \citep{DBLP:conf/nips/TirumalaSAM23} is adopted as the parameter-updated objective.

\subsection{DVLM-based OPD}
\label{sec: DPO}

\paragraph{Overview.} 
For a trained DVLM teacher, we can directly use it as a new teacher model to supervise the student during OPD. 
Specifically, we likewise feed the PTS into the teacher model and use HPA to control the information flow. 
We then extract the logits at the last token position of each re-tokenized group as the supervision signal for the corresponding student token. 
% Hence, DVLM is aligned with the student model to be trained in both the tokenization scheme and the vocabulary space.

Given a prompt $c$, the student model $p_\theta(\cdot\mid\cdot)$ first generates an on-policy response $\mathbf{y}$ with length $L_R$.
$\theta$ denotes the parameters of the student model.
\begin{equation}
\mathbf{y}=(y_1,\ldots,y_{L_R})
\sim p_\theta(\cdot\mid c),
\end{equation}

We then set $x=[c;\mathbf{y}]$ and construct $\mathbf{X}_{\mathrm{PTS}}$ following the same procedure as Equation~\ref{eq: pts}.
The resulting $\mathbf{X}_{\mathrm{PTS}}$ is fed into the frozen DVLM teacher with HPA (Equation~\ref{eq:4}). 
For each response token $y_i$, we extract its $\mathbf{h}^{e}_{i,m_i}$ from the last position of its re-tokenized group. 
The well-trained student-vocabulary head then produces the corresponding DVLM teacher logits:
\begin{equation}
\mathbf{logit}_i^D
=
\mathbf{W}_D^T \mathbf{h}^{e}_{i,m_i}
\in\mathbb{R}^{|V_S|},
\qquad i=1,\ldots,L_R.
\end{equation}
where $\mathbf{logit}_i^D$ means the $i$-th valid supervision logit vector produced by the DVLM teacher, corresponding to the $i$-th student-generated token in $\mathbf{y}$.

The DVLM teacher and the extraction strategy of PTS hidden states jointly ensure that both the input tokens and the output logit dimensionalities (that is, the distribution) are aligned with those of the student model to be trained. 
Consequently, the teacher distribution shape is transformed from $L_T \times |V_T|$ of the original teacher model to $L_S \times |V_S|$.
We can then directly optimize the student model using forward KL divergence \citep{hinton2015distillingknowledgeneuralnetwork,gu2026minillmonpolicydistillationlarge}.
\begin{equation}
\mathcal{L}
=
\frac{1}{L_R}\sum_{i=1}^{L_R}
D_{\mathrm{KL}}\!\left(
\operatorname{softmax}(\mathbf{logit}_i^D)
\,\middle\|\,
\operatorname{softmax}(\mathbf{logit}_i^S)
\right).
\end{equation}
where \(D_{\mathrm{KL}}\) denotes the KL divergence, `\(\operatorname{softmax}\)' converts logits into probability distributions, and \(\mathbf{logit}_i^S\) represents the student logits at the \(i\)-th response position.

\section{Experiments}
\label{sec: exp}
In this section, we provide a detailed description of the training configurations and evaluation protocols in our experiments. 
We then conduct comprehensive assessments from multiple perspectives to demonstrate the effectiveness of our
approach.
In particular, the term `DVLM' refers to our newly proposed teacher-model architecture. 
It can denote either the teacher model itself (we refer to it as: `the DVLM teacher') or the overall method (we refer to it as: `the method DVLM').
We distinguish between the two usages where necessary in the following parts.

\subsection{Experimental Settings}
\label{sec: experimental settings}
\subsubsection{Training}
\label{sec: training settings}
\paragraph{Models.}
In this work, we use Qwen3-4B \citep{yang2025qwen3technicalreport} as the teacher model, Llama-3.2-1B-Instruct \citep{DBLP:journals/corr/abs-2407-21783} and OLMo-2-1B-Instruct \citep{olmo20252olmo2furious} as the student models.
The vocabulary overlap ratios of the two student models with the teacher model are around 83.8\% and 59.5\%, respectively. 
This substantial difference makes them representative test models.

\paragraph{Data.}
For the training of the new LM heads, we use the commonly adopted pretraining corpus Nemotron-CC-v2 \citep{nvidia2025nvidianemotronnano2} (approximately $0.2$B tokens), augmented with the LLM reasoning dataset Skywork-OR1-Data \citep{he2025skyworkopenreasoner1} (approximately $0.02$B tokens), for a total of $0.22$B tokens.
The primary purpose of mixing these datasets is to expose the new LM heads to as many diverse tokens as possible.
For DVLM-based OPD, we follow \citet{wang2026crosstokenizeronpolicydistillationbyteprefix} and combine DAPO-Math \citep{yu2025dapoopensourcellmreinforcement} (math-related) and TACO \citep{li2023tacotopicsalgorithmiccode} (code-related) training data to construct the training set.
For DAPO-Math, $32$ rollouts are sampled from the teacher model for each problem, and we retain problems with a pass rate of at least $25$\%. 
For TACO, we select problems labeled as `EASY' or `MEDIUM'. 
This filtering retains problems that are moderately challenging without being overly difficult, thereby ensuring the quality of student trajectories and the effectiveness of teacher supervision.
The final OPD training set contains around $8$K samples.

% \paragraph{Training Implementation.}
% All experiments are conducted on $4$ NVIDIA H$200$ GPUs, each with $141$ GB of memory. 
% For [S1], we use only one GPU. 
% For [S2], we deploy the student model with vLLM \citep{kwon2023efficient} on one GPU to generate responses for OPD and refresh its weights as the student parameters are updated. 
% The remaining three GPUs are used for data-parallel training.
% The code for [S2] is adapted from HuggingFace’s TRL \citep{vonwerra2020trl} library.

\subsubsection{Evaluation}
\label{sec: evaluation settings}

\paragraph{Benchmarks.}
To evaluate the trained DVLM teacher after Stage I, we select the following four language-understanding benchmarks:
BoolQ \citep{clark-etal-2019-boolq}, WinoGrande \citep{sakaguchi2019winograndeadversarialwinogradschema}, Arc-Challenge \citep{clark2018thinksolvedquestionanswering} and MMLU(Algebra) \citep{hendrycks2021measuringmassivemultitasklanguage}.
For the distillation experiments in Stage II, we evaluate the student model on three mathematical reasoning benchmarks: GSM8K \citep{cobbe2021trainingverifierssolvemath}, MATH-500 \citep{lightman2023letsverifystepstep}, and AMC23 \citep{yang2024qwen25mathtechnicalreportmathematical}, along with three code benchmarks: HumanEval+ \citep{liu2023codegeneratedchatgptreally}, CRUXEval \citep{gu2024cruxevalbenchmarkcodereasoning}, and LiveCodeBench v5 \citep{jain2024livecodebenchholisticcontaminationfree}.

\paragraph{Baselines.} We consider two categories of baselines: common LLM reasoning enhancement methods \citep{xu2025largereasoningmodelssurvey,chen2026surveyinductivereasoninglarge}, including supervised fine-tuning (SFT) \citep{zhang2025instructiontuninglargelanguage} and reinforcement learning (RL) \citep{zhang2025surveyreinforcementlearninglarge}, specifically GRPO \citep{shao2024deepseekmathpushinglimitsmathematical}; and four types of cross-tokenizer distillation methods: DSKD (representation-based) \citep{zhang2024dualspaceknowledgedistillationlarge}, ALM (span-based) \citep{minixhofer2025universalcrosstokenizerdistillationapproximate}, GOLD (from which our code is adapted) \citep{patiño2025_unlocking_on_policy_distillation_for_any_model_family}, and BPM (byte-based) \citep{wang2026crosstokenizeronpolicydistillationbyteprefix}.

\begin{table}[t]
\centering
\captionsetup{skip=1.5pt}
\caption{Main results of our cross-tokenizer distillation method on six reasoning benchmarks with two student models. The best results are highlighted in dark yellow and bold, while the second-best cross-tokenizer distillation results are marked in light yellow. `$\Delta$ w/ base’ denotes the improvement of our method over the student base model. Values in $()$ indicate the variation across multiple runs.}
\fontsize{9pt}{9.5pt}\selectfont
\setlength{\tabcolsep}{3.0pt}
\renewcommand{\arraystretch}{1.2}
\begin{tabularx}{\linewidth}{
>{\raggedright\arraybackslash}p{1.8cm}
*{7}{>{\centering\arraybackslash}X}
}
\toprule
\multirow{3}{*}{{\fontsize{10pt}{12pt}\selectfont Method}} & \multicolumn{3}{c}{{\fontsize{10pt}{12pt}\selectfont Mathematics}} & \multicolumn{3}{c}{{\fontsize{10pt}{12pt}\selectfont Coding}} & \multirow{3}{*}{{\fontsize{10pt}{12pt}\selectfont Average}} \\
\cmidrule(lr){2-4} \cmidrule(lr){5-7}
& GSM8K & MATH-500 & AMC23 & HumanEval+ & CRUXEval & \makecell{LCBench} & \\
& ACC. & ACC. & pass@1 & pass@1 & pass@1 & pass@1 & \\
\midrule
\rowcolor{TeacherBlue}
Qwen3-4B & $87.79$ & $68.40$ & $38.28$ & $82.93$ & $41.81$ & $26.93$ & $57.69$ \\
\midrule
\rowcolor{HeaderOrange}
Llama-3.2-1B & $44.05$ & $27.00$ & $4.68$ & $31.10$ & $14.25$ & $1.47$ & $20.43$ \\
SFT
& $45.06_{\scriptscriptstyle(0.4)}$
& $26.47_{\scriptscriptstyle(0.3)}$
& $4.84_{\scriptscriptstyle(0.3)}$
& $33.13_{\scriptscriptstyle(0.9)}$
& $8.38_{\scriptscriptstyle(0.4)}$
& $1.61_{\scriptscriptstyle(0.2)}$
& $19.92$ \\
GRPO
& $43.34_{\scriptscriptstyle(0.2)}$
& $27.00_{\scriptscriptstyle(0.4)}$
& $6.71_{\scriptscriptstyle(0.2)}$
& $31.30_{\scriptscriptstyle(0.9)}$
& $14.42_{\scriptscriptstyle(0.5)}$
& $1.93_{\scriptscriptstyle(0.1)}$
& $20.78$ \\
\mydashline
DSKD
& $42.51_{\scriptscriptstyle(0.3)}$
& $23.93_{\scriptscriptstyle(0.4)}$
& $5.52_{\scriptscriptstyle(0.3)}$
& $31.91_{\scriptscriptstyle(0.9)}$
& $13.85_{\scriptscriptstyle(0.3)}$
& $1.66_{\scriptscriptstyle(0.3)}$
& $19.90$ \\
ALM
& $43.32_{\scriptscriptstyle(0.2)}$
& $25.73_{\scriptscriptstyle(0.4)}$
& $5.83_{\scriptscriptstyle(0.4)}$
& \second{$34.35_{\scriptscriptstyle(0.9)}$}
& $14.60_{\scriptscriptstyle(0.3)}$
& $1.72_{\scriptscriptstyle(0.1)}$
& $20.93$ \\
GOLD
& $43.04_{\scriptscriptstyle(0.2)}$
& \second{$27.27_{\scriptscriptstyle(0.4)}$}
& \second{$5.93_{\scriptscriptstyle(0.4)}$}
& \second{$34.35_{\scriptscriptstyle(0.9)}$}
& $13.94_{\scriptscriptstyle(0.2)}$
& $1.48_{\scriptscriptstyle(0.1)}$
& $21.00$ \\
BPM
& \second{$44.25_{\scriptscriptstyle(0.2)}$}
& $26.73_{\scriptscriptstyle(0.4)}$
& $5.78_{\scriptscriptstyle(0.6)}$
& $33.94_{\scriptscriptstyle(0.9)}$
& \second{$15.73_{\scriptscriptstyle(0.2)}$}
& \second{$1.79_{\scriptscriptstyle(0.5)}$}
& \second{$21.37$} \\
\textbf{DVLM}
& \best{$45.89$}$_{\scriptscriptstyle(0.2)}$
& \best{$29.93$}$_{\scriptscriptstyle(0.3)}$
& \best{$9.06$}$_{\scriptscriptstyle(0.3)}$
& \best{$36.99$}$_{\scriptscriptstyle(0.9)}$
& \best{$16.44$}$_{\scriptscriptstyle(0.4)}$
& \best{$2.86$}$_{\scriptscriptstyle(0.1)}$
& \best{$23.53$} \\
$\Delta$ w/ base
& \textcolor{red}{$+1.84$}
& \textcolor{red}{$+2.93$}
& \textcolor{red}{$+4.38$}
& \textcolor{red}{$+5.89$}
& \textcolor{red}{$+2.19$}
& \textcolor{red}{$+1.39$}
& \textcolor{red}{$+3.10$} \\
\midrule
\rowcolor{HeaderOrange}
OLMo-2-1B & $69.37$ & $21.00$ & $5.00$ & $23.17$ & $12.31$ & $0.27$ & $21.85$ \\
SFT
& $63.89_{\scriptscriptstyle(0.5)}$
& $17.07_{\scriptscriptstyle(0.4)}$
& $3.12_{\scriptscriptstyle(0.5)}$
& $19.92_{\scriptscriptstyle(0.9)}$
& $15.12_{\scriptscriptstyle(0.4)}$
& $0.31_{\scriptscriptstyle(0.4)}$
& $19.91$ \\
GRPO
& $69.80_{\scriptscriptstyle(0.4)}$
& $21.80_{\scriptscriptstyle(0.4)}$
& $5.93_{\scriptscriptstyle(0.3)}$
& $21.54_{\scriptscriptstyle(0.9)}$
& $12.25_{\scriptscriptstyle(0.5)}$
& $0.22_{\scriptscriptstyle(0.2)}$
& $21.92$ \\
\mydashline
DSKD
& $64.42_{\scriptscriptstyle(0.2)}$
& $17.93_{\scriptscriptstyle(0.4)}$
& $5.88_{\scriptscriptstyle(0.5)}$
& $18.50_{\scriptscriptstyle(0.9)}$
& \second{$15.25_{\scriptscriptstyle(0.4)}$}
& $0.10_{\scriptscriptstyle(0.5)}$
& $20.35$ \\
ALM
& $66.44_{\scriptscriptstyle(0.4)}$
& \second{$21.07_{\scriptscriptstyle(0.4)}$}
& \second{$6.35_{\scriptscriptstyle(0.4)}$}
& $23.98_{\scriptscriptstyle(0.9)}$
& $14.75_{\scriptscriptstyle(0.5)}$
& $0.23_{\scriptscriptstyle(0.2)}$
& $22.14$ \\
GOLD
& $67.27_{\scriptscriptstyle(0.3)}$
& $19.27_{\scriptscriptstyle(0.3)}$
& $5.26_{\scriptscriptstyle(0.5)}$
& \second{$25.20_{\scriptscriptstyle(0.9)}$}
& $13.54_{\scriptscriptstyle(0.5)}$
& $0.45_{\scriptscriptstyle(0.3)}$
& $21.83$ \\
BPM
& \second{$69.35_{\scriptscriptstyle(0.5)}$}
& \second{$21.07_{\scriptscriptstyle(0.4)}$}
& $6.15_{\scriptscriptstyle(0.4)}$
& $24.39_{\scriptscriptstyle(0.7)}$
& $15.15_{\scriptscriptstyle(0.3)}$
& \second{$0.88_{\scriptscriptstyle(0.4)}$}
& \second{$22.83$} \\
\textbf{DVLM}
& \best{$70.15$}$_{\scriptscriptstyle(0.2)}$
& \best{$22.87$}$_{\scriptscriptstyle(0.4)}$
& \best{$8.39$}$_{\scriptscriptstyle(0.3)}$
& \best{$27.44$}$_{\scriptscriptstyle(0.7)}$
& \best{$17.94$}$_{\scriptscriptstyle(0.4)}$
& \best{$1.50$}$_{\scriptscriptstyle(0.2)}$
& \best{$24.72$} \\
$\Delta$ w/ base
& \textcolor{red}{$+0.78$}
& \textcolor{red}{$+1.87$}
& \textcolor{red}{$+3.39$}
& \textcolor{red}{$+4.27$}
& \textcolor{red}{$+5.63$}
& \textcolor{red}{$+1.23$}
& \textcolor{red}{$+2.86$} \\
\bottomrule
\end{tabularx}
\label{tab:main_results}
\end{table}

\subsection{Main Results}
Table~\ref{tab:main_results} presents the results of our main experiments, from which we draw the following conclusions.
\begin{enumerate}[label=(\arabic*), leftmargin=2em, itemsep=2pt, topsep=2pt, parsep=0pt, partopsep=0pt]
    \item \textbf{Our method DVLM is more effective for cross-tokenizer distillation.}
    Despite being evaluated in a setting for generalization, our method improves the two models by $3.10$ and $2.86$ points, respectively, across the six reasoning benchmarks. 
    Compared with the second-best cross-tokenizer OPD method, our method DVLM achieves further gains of $2.16$ and $1.89$ points separately. 
    Relative to the student base models, our method also yields average improvements of $2.53$ on the math domain and $3.43$ on the code domain.
    
    \item \textbf{Beyond SFT and RL, cross-tokenizer OPD shows great potential.}
    Compared with conventional LLM reasoning enhancement methods such as SFT and RL, even cross-tokenizer OPD can demonstrate strong behaviors, suggesting its promising potential for LLM training. 
    By addressing the core challenges of cross-tokenizer settings, our method DVLM combines the advantages of both and therefore achieves more substantial performance gains.

    \item \textbf{Constructing dimension-aligned distributions is the key in our method.}
    There is a huge performance gap between the base students and the teacher, while the training and evaluation also follow a challenging generalization setting (Section~\ref{sec: training settings}). 
    Only our proposed method, DVLM, constructs the dimensionally aligned supervision that exactly matches the student’s original distribution, enabling effective distillation on challenging problems.

    \item \textbf{Our method DVLM is especially effective for student models that have large vocabulary gaps with teacher models.}  
    For OLMo-2-1B, whose vocabulary differs more substantially from that of the teacher Qwen3-4B, methods such as ALM, GOLD, and BPM can improve average performance but do not guarantee consistent gains across every benchmark. 
    Their span-level or byte-level alternatives bypass token misalignment but disrupt the model’s native tokenization and pretraining paradigm, whereas our method DVLM addresses this issue at its root.
\end{enumerate}

\subsection{Ablation Study}
\label{sec: abla}
With settings following those applied in the main experiments, we conduct ablation studies on both student models to validate the effectiveness of each component of our method.
The results are in Table~\ref{tab:ablation}, and we can draw the following conclusions.
\begin{enumerate}[label=(\arabic*), leftmargin=2em, itemsep=2pt, topsep=2pt, parsep=0pt, partopsep=0pt]
    \item \textbf{Every part in our method DVLM matters.}
    As clearly shown in the table, every component of our method contributes positively to the final performance and is therefore indispensable, as removing any component leads to a performance drop for both student models.
    
    \item \textbf{The quality of the DVLM teacher is the core.} 
    The training quality of the new LM head in the DVLM teacher plays the most critical role in cross-tokenizer distillation (`w/o HPA', `w/o T.Prefix', and `w/o Re-token.'). 
    Among these components, the construction of PTS is particularly important, as the teacher input should preserve as many native teacher tokens as possible to avoid train–inference mismatch while learning the weights of student tokens. 
    Our HPA mechanism prevents information leakage and further ensures this property.
    
    \item \textbf{The LM head and logit extraction make the distributional dimensions match the student.} 
    Once the new LM head is trained, it can effectively support student-output logits, as removing it (`w/o LM Head') causes a certain degree of performance degradation. 
    Meanwhile, to fully exploit the capability of our teacher, extracting the last-position logit of each re-tokenized group, forming a student-like tokenization schema, is also important (`w/o L.Extract').
\end{enumerate}

\begin{table}[t]
\centering
\captionsetup{skip=1.5pt}
\caption{Results of ablation studies. `w/o HPA’, `w/o T.Prefix', `w/o Re-token.', `w/o LM Head', and `w/o L.Extract' respectively mean standard causal attention, the DVLM teacher input with only re-tokenized tokens, the DVLM teacher input with only the teacher sequence, OPD with the original teacher model, and utilizing span-level distillation to replace logit extraction in our method.}
\fontsize{9pt}{9.5pt}\selectfont
\setlength{\tabcolsep}{3.0pt}
\renewcommand{\arraystretch}{1.2}
\begin{tabularx}{\linewidth}{
>{\raggedright\arraybackslash}p{1.8cm}
*{7}{>{\centering\arraybackslash}X}
}
\toprule
\multirow{3}{*}{{\fontsize{10pt}{12pt}\selectfont Method}} & \multicolumn{3}{c}{{\fontsize{10pt}{12pt}\selectfont Mathematics}} & \multicolumn{3}{c}{{\fontsize{10pt}{12pt}\selectfont Coding}} & \multirow{3}{*}{{\fontsize{10pt}{12pt}\selectfont Average}} \\
\cmidrule(lr){2-4} \cmidrule(lr){5-7}
& GSM8K & MATH-500 & AMC23 & HumanEval+ & CRUXEval & \makecell{LCBench} & \\
& ACC. & ACC. & pass@1 & pass@1 & pass@1 & pass@1 &  \\
\midrule
\rowcolor{HeaderOrange}
Llama-3.2-1B & $44.05$ & $27.00$ & $4.68$ & $31.10$ & $14.25$ & $1.47$ & $20.43$ \\
\textbf{DVLM} & \best{$45.89$} & \best{$29.93$} & \best{$9.06$} & \best{$36.99$} & \best{$16.44$} & \best{$2.86$} & \best{$23.53$} \\
\mydashline
w/o HPA & $38.36$ & $18.40$ & $2.58$ & $23.17$ & $12.43$ & $1.02$ & $15.99$\\
w/o T.Prefix & $38.28$ & $17.00$ & $2.33$ & $21.34$ & $11.81$ & $1.32$ & $15.35$\\
w/o Re-token. & $37.91$ & $16.00$ & $2.16$ & $18.29$ & $11.43$ & $0.88$ & $14.45$\\
w/o LM Head & $43.21$ & $25.60$ & $3.34$ & $31.70$ & $14.37$ & $1.48$ & $19.95$\\
w/o L.Extract & $44.73$ & $27.00$ & $4.43$ & $35.36$ & $14.87$ & $2.05$ & $21.41$\\
\midrule
\rowcolor{HeaderOrange}
OLMo-2-1B & $69.37$ & $21.00$ & $5.00$ & $23.17$ & $12.31$ & $0.27$ & $21.85$ \\
\textbf{DVLM} & \best{$70.15$} & \best{$22.87$} & \best{$8.39$} & \best{$27.44$} & \best{$17.94$} & \best{$1.50$} & \best{$24.72$} \\
\mydashline
w/o HPA & $64.44$ & $9.80$ & $1.12$ & $15.24$ & $11.06$ & $0.05$ & $16.95$\\
w/o T.Prefix & $61.41$ & $10.00$ & $1.08$ & $12.19$ & $10.05$ & $0.00$ & $15.79$\\
w/o Re-token. & $51.54$ & $8.80$ & $0.95$ & $9.14$ & $9.75$ & $0.00$ & $13.36$\\
w/o LM Head & $68.23$ & $20.20$ & $2.17$ & $20.12$ & $11.75$ & $0.70$ & $20.53$\\
w/o L.Extract & $69.74$ & $21.20$ & $3.60$ & $20.73$ & $12.12$ & $0.64$ & $21.34$\\
\bottomrule
\end{tabularx}
\label{tab:ablation}
\end{table}

\subsection{In-domain Evaluation}
\label{sec: in-domain}
\begin{figure*}[t]
    \centering

    \begin{subfigure}[t]{0.24\textwidth}
        \centering
        \includegraphics[width=\linewidth]{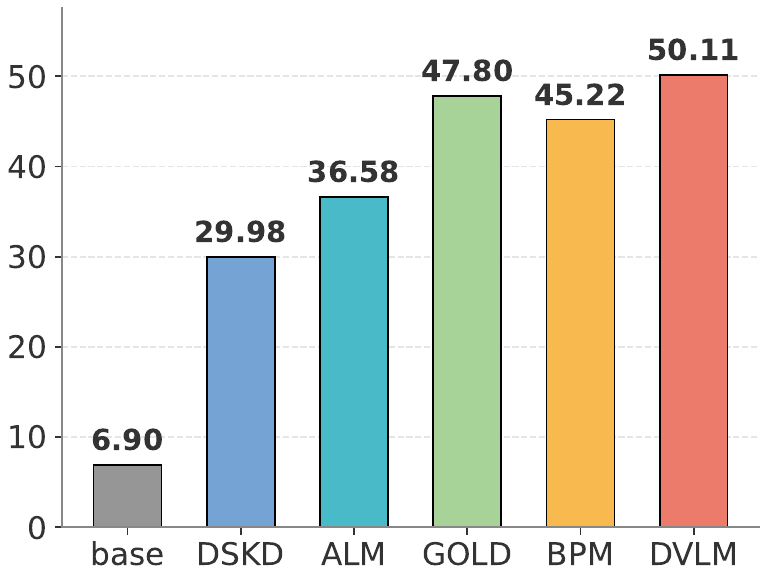}
        \caption{Llama on Countdown.}
    \end{subfigure}%
    \hfill
    \begin{subfigure}[t]{0.24\textwidth}
        \centering
        \includegraphics[width=\linewidth]{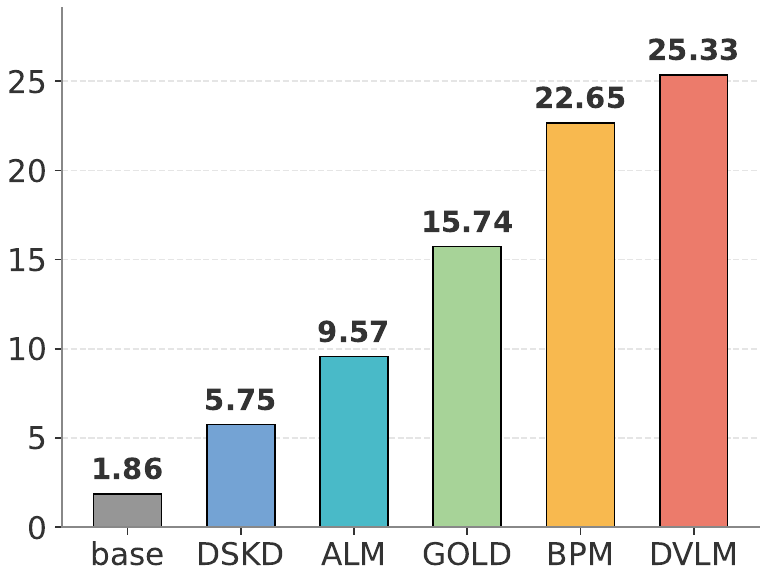}
        \caption{OLMo on Countdown.}
    \end{subfigure}%
    \hfill
    \begin{subfigure}[t]{0.24\textwidth}
        \centering
        \includegraphics[width=\linewidth]{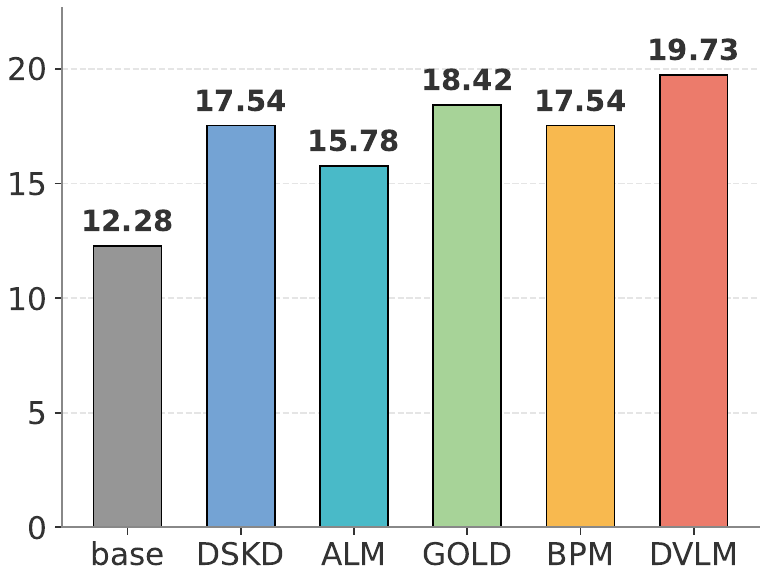}
        \caption{Llama on LeetCode.}
    \end{subfigure}%
    \hfill
    \begin{subfigure}[t]{0.24\textwidth}
        \centering
        \includegraphics[width=\linewidth]{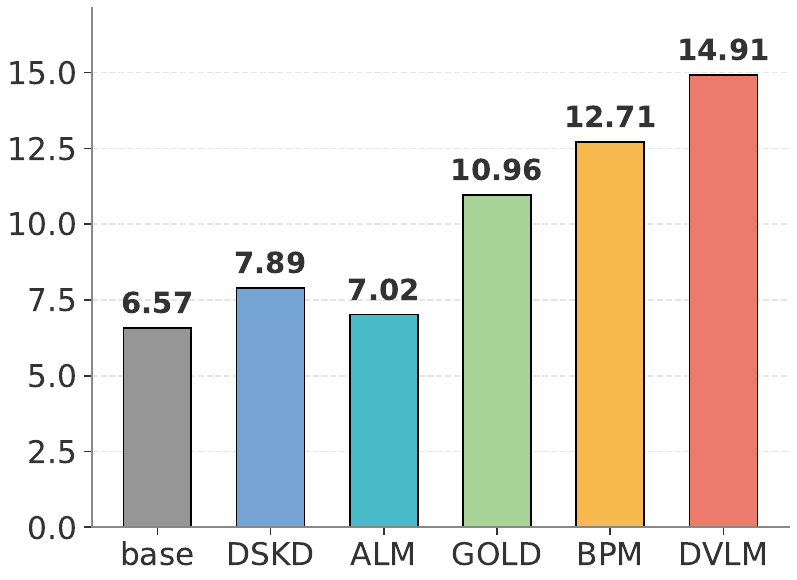}
        \caption{OLMo on LeetCode.}
    \end{subfigure}

    \caption{Results of in-domain evaluation of two student models on two datasets.}
    \label{fig:4}
\end{figure*}

Since our main experiments focus on generalization, we further validate the effectiveness of our method DVLM, through in-domain evaluation. 
Specifically, we select Countdown \citep{gandhi2024streamsearchsoslearning} and LeetCodeDataset \citep{xia2025leetcodedatasettemporaldatasetrobust} for the mathematics and code domains, respectively, and train both student models on the training sets before evaluating them on the corresponding test sets.
Consistent with the baselines in Table~\ref{tab:main_results}, we compare our method DVLM with four other cross-tokenizer distillation methods. 
The results in Figure~\ref{fig:4} show that ours achieves the best distillation performance for both student models, Llama-3.2-1B and OLMo-2-1B, on both datasets.
On Countdown, our method DVLM outperforms the strongest baselines by $2.31$ and $2.68$ points, respectively. 
On LeetCodeDataset, it exceeds the best competing baselines by $1.31$ and $2.20$ points. 
We attribute OLMo’s lower scores compared with Llama primarily to its substantially larger vocabulary discrepancy with the teacher, which makes cross-tokenizer distillation considerably more challenging.
Nevertheless, our method still achieves the best performance.

\subsection{Evaluation of the DVLM Architecture}
\begin{figure*}[t]
    \centering

    \begin{subfigure}[t]{0.48\textwidth}
        \centering
        \includegraphics[width=\linewidth]{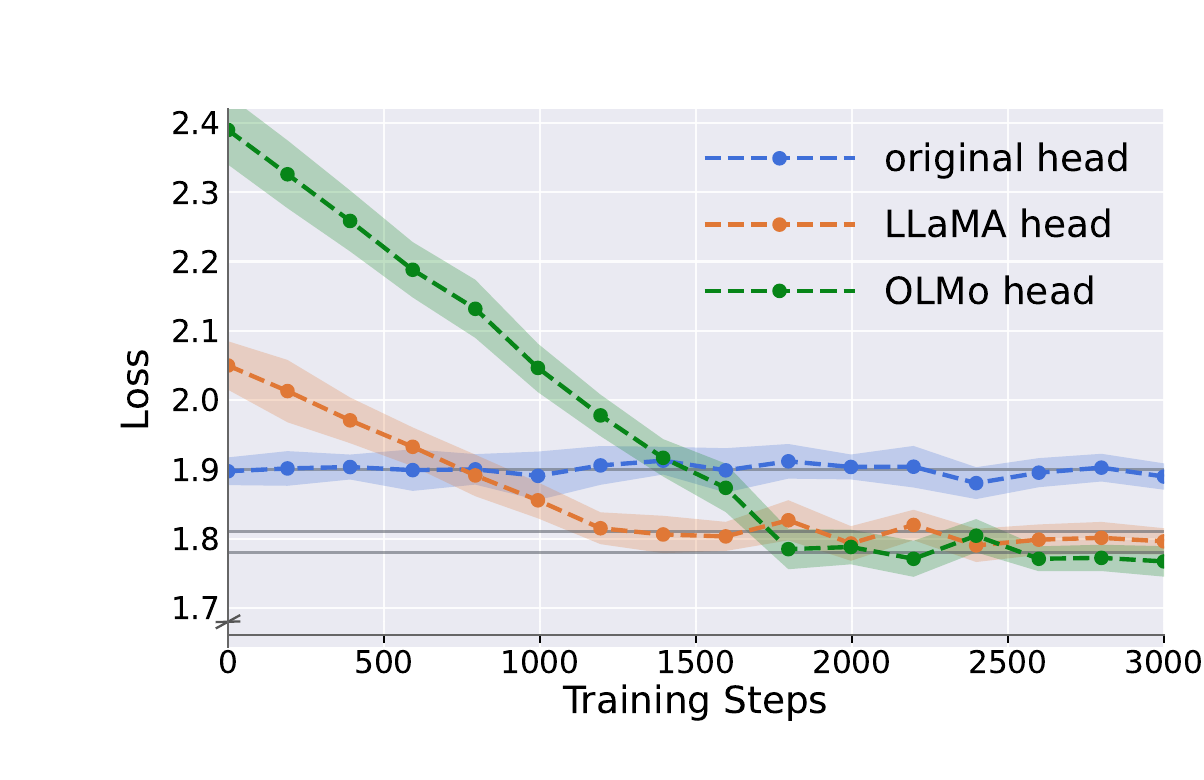}
        \caption{Loss curves of different LM heads.}
        \label{fig3:sub_a}
    \end{subfigure}%
    % \hspace{0.01\textwidth}%
    \begin{subfigure}[t]{0.5\textwidth}
        \centering
        \includegraphics[width=\linewidth]{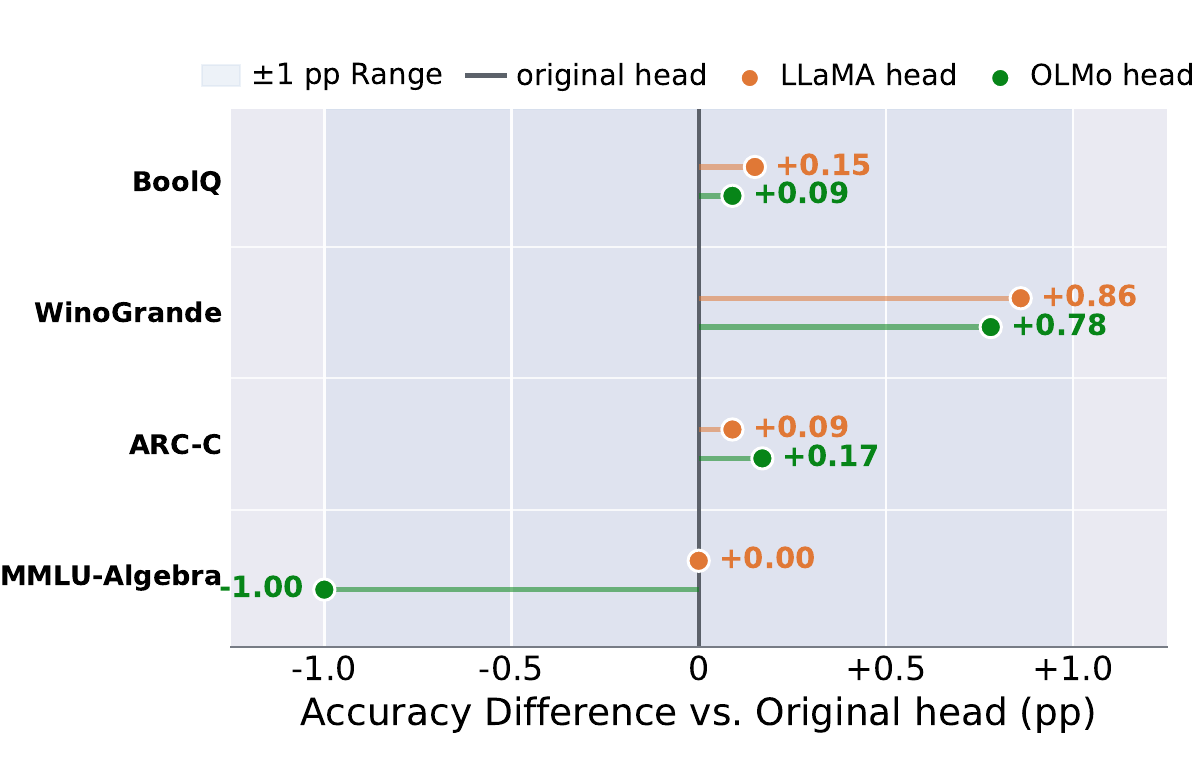}
        \caption{Understanding comparison with the original teacher.}
        \label{fig3:sub_b}
    \end{subfigure}

    \caption{The effectiveness of the DVLM teacher is evaluated by comparing both the training loss (the shaded areas represent the error ranges across multiple runs) and the performances on four understanding tasks against those of the original teacher model.}
    \label{fig:3}
\end{figure*}

All the above evaluations focus on cross-vocabulary distillation in Stage II. 
We also test the new teacher, DVLM, trained in Stage I.
Since training the new LM head closely resembles pretraining, we measure it from two perspectives: loss convergence and the performance of the new model architecture on LLM understanding tasks.
We follow the experimental setup described in Section~\ref{sec: experimental settings}.
Figure~\ref{fig3:sub_a} shows that the loss of the original teacher model stays approximately $1.9$, while the teacher losses with the Llama and OLMo heads decreased to $1.81$ at around step $1,200$ and $1.79$ at around step $1,750$, respectively.
This indicates that our training method achieves effective convergence.
The loss curves further demonstrate that larger vocabulary gaps lead to a higher initial loss but lower final loss values.
Figure~\ref{fig3:sub_b} further indicates that, although the OLMo-head model performs $1$ point lower on MMLU(Algebra), both DVLM teachers can overall match the teacher in text understanding performance.
This $1$ point difference is mainly due to the small sample size of the MMLU(Algebra) subset and is therefore within a reasonable range.
More experiments are in Appendix~\ref{app: dvlm per}.

\subsection{Difficulty Selection for Cross-tokenizer Distillation}
\label{sec: diff}
\begin{figure*}[t]
    \centering

    \begin{subfigure}[t]{0.24\textwidth}
        \centering
        \includegraphics[width=\linewidth]{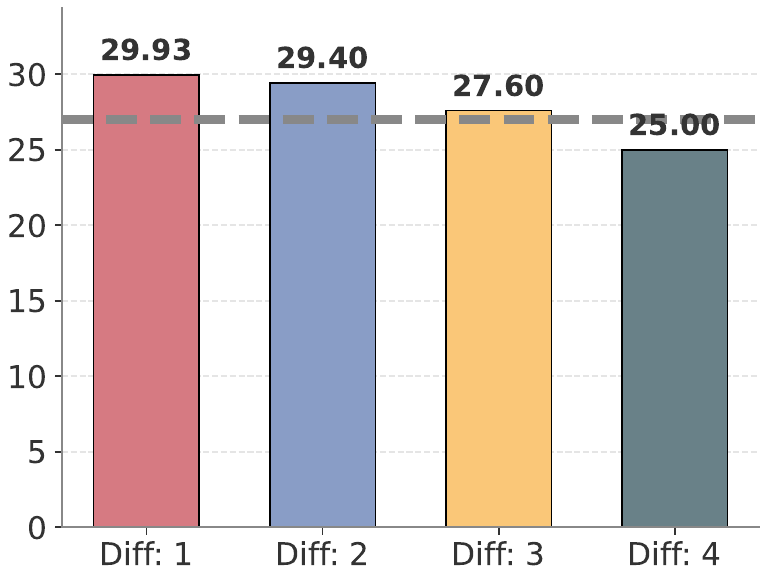}
        \caption{Results on Math-500.}
    \end{subfigure}%
    \hfill
    \begin{subfigure}[t]{0.24\textwidth}
        \centering
        \includegraphics[width=\linewidth]{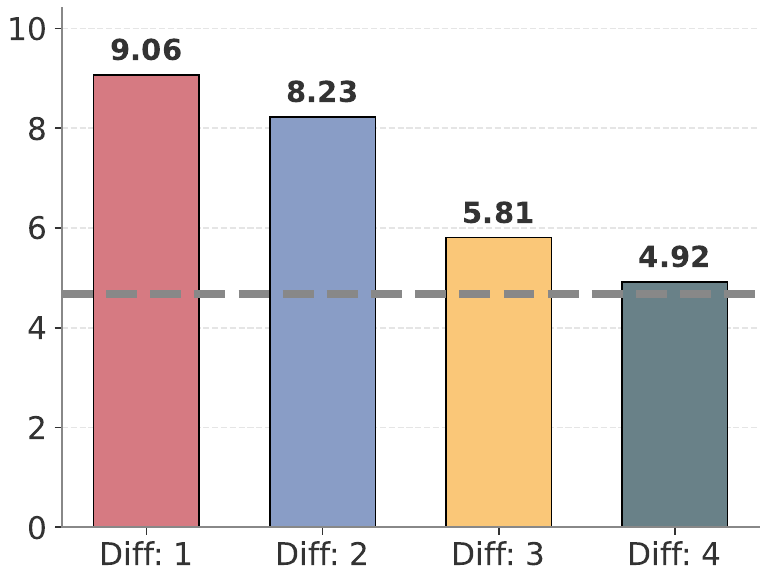}
        \caption{Results on AMC23.}
    \end{subfigure}%
    \hfill
    \begin{subfigure}[t]{0.24\textwidth}
        \centering
        \includegraphics[width=\linewidth]{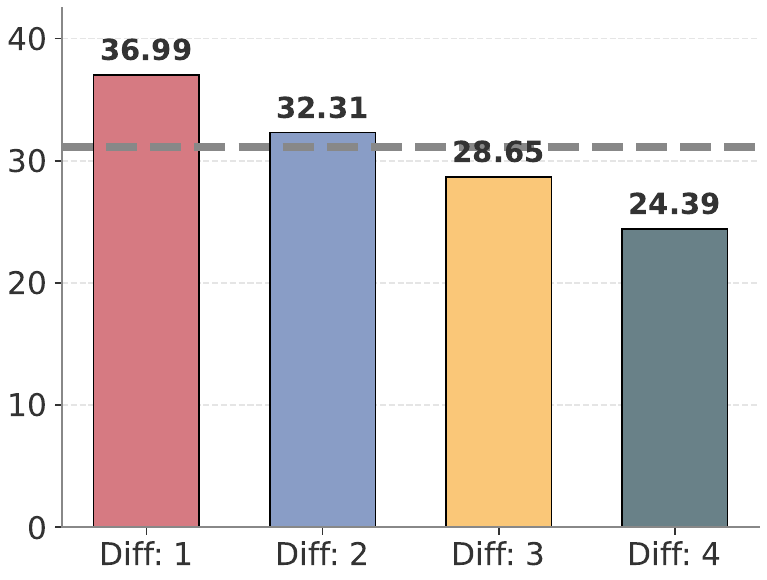}
        \caption{Results on Heval+.}
    \end{subfigure}%
    \hfill
    \begin{subfigure}[t]{0.24\textwidth}
        \centering
        \includegraphics[width=\linewidth]{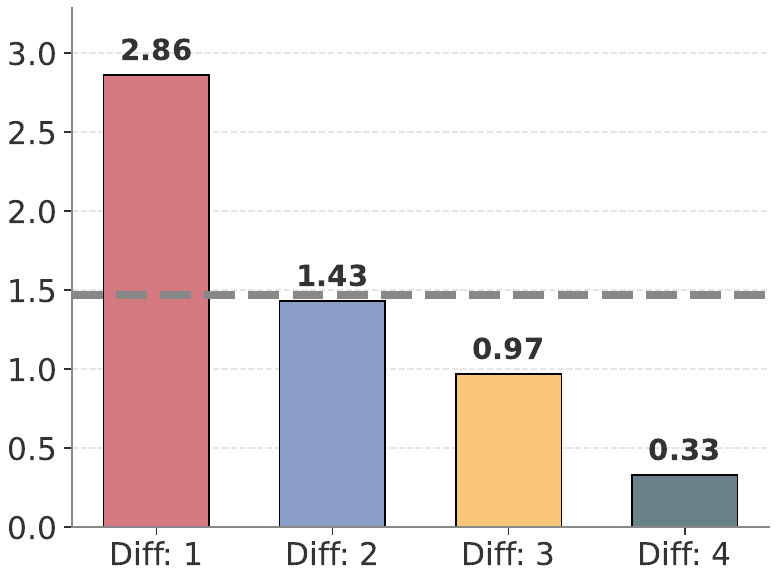}
        \caption{Results on LCBench.}
    \end{subfigure}

    \caption{Sample difficulty scaling on four benchmarks with Llama-3.2-1B.}
    \label{fig:5}
\end{figure*}

We further investigate the impact of sample difficulty \citep{havrilla2024surveyingeffectsqualitydiversity} on training performance. 
We designate the difficulty of our original dataset as `Diff: 1'. 
For TACO(`HARD'), we also perform $32$ rollouts with the teacher to assess the difficulty. 
We then progressively increase sample difficulty while keeping the number of training samples and the math-code ratio fixed, constructing three increasingly difficult datasets: `Diff: 2', `Diff: 3', and `Diff: 4'.
The results in Figure~\ref{fig:5} show that increasingly difficult samples are less likely to yield effective distillation and may even degrade the performance of the original backbone (below the gray lines). 
These findings suggest that, in cross-tokenizer distillation, sample difficulty should match student capability, as overly difficult samples will make the student produce poor trajectories and degrade performance.

\section{Conclusion}
In this paper, we aim to address the challenges in cross-tokenizer distillation.
We propose the Dual-Vocabulary Language Model, which enables the teacher model to produce logits over the student vocabulary by taking a Parallel-Tokenized Sequence as input and controlling information flow with Hybrid-Prefix Attention. 
DVLM-based OPD extracts the teacher distribution in the same manner, with logit dimensions matching those of the student model.
Experiments prove that our method consistently improves the student models across all six reasoning benchmarks.

\section*{AI use statement}

In this work, we used generative AI tools for coding assistance, polishing and improving the writing of the manuscript, and generating example images used in figures. 
We have reviewed all AI-assisted work. AI-assisted code was manually inspected and tested for correctness, manuscript revisions were carefully reviewed by the authors to ensure accuracy and consistency with the intended meaning, and AI-generated content was checked and selected by the authors before inclusion. 
We take responsibility for the final content of this work, including text, claims, code, and artifacts produced with the aid of generative AI.

\section*{Ethics statement}

All models and datasets used in this work are obtained from open-source resources released by the academic community. 
All training and LLM evaluation frameworks used in this work are publicly available on GitHub, and our code is adapted from these open-source frameworks.
We do not identify any concerns related to content misuse, conflicts of interest, bias or fairness, privacy or security, legal compliance, or research integrity.

\section*{Reproducibility statement}
During the submission period, we provide all code and training data through an anonymous repository and the supplementary materials. These resources will be made publicly available after the submission process is completed.
Detailed descriptions of the model architecture, training procedure, data processing pipeline, hyperparameter settings, and evaluation protocols are provided in the main paper and appendix.

\section*{Acknowledgments}
We will include this content in the camera-ready version.

\bibliography{iclr2027_conference}

\begin{thebibliography}{62}
\providecommand{\natexlab}[1]{#1}
\providecommand{\url}[1]{\texttt{#1}}
\expandafter\ifx\csname urlstyle\endcsname\relax
  \providecommand{\doi}[1]{doi: #1}\else
  \providecommand{\doi}{doi: \begingroup \urlstyle{rm}\Url}\fi

\bibitem[Agarwal et~al.(2024)Agarwal, Vieillard, Zhou, Stanczyk, Ramos, Geist, and Bachem]{agarwal2024onpolicydistillationlanguagemodels}
Rishabh Agarwal, Nino Vieillard, Yongchao Zhou, Piotr Stanczyk, Sabela Ramos, Matthieu Geist, and Olivier Bachem.
\newblock On-policy distillation of language models: Learning from self-generated mistakes, 2024.
\newblock URL \url{https://arxiv.org/abs/2306.13649}.

\bibitem[Ali et~al.(2024)Ali, Fromm, Thellmann, Rutmann, L{\"{u}}bbering, Leveling, Klug, Ebert, Doll, Buschhoff, Jain, Weber, Jurkschat, Abdelwahab, John, Suarez, Ostendorff, Weinbach, Sifa, Kesselheim, and Flores{-}Herr]{DBLP:conf/naacl/AliFTRLLKEDBJWJAJSOWSKF24}
Mehdi Ali, Michael Fromm, Klaudia Thellmann, Richard Rutmann, Max L{\"{u}}bbering, Johannes Leveling, Katrin Klug, Jan Ebert, Niclas Doll, Jasper~Schulze Buschhoff, Charvi Jain, Alexander~Arno Weber, Lena Jurkschat, Hammam Abdelwahab, Chelsea~Maria John, Pedro~Ortiz Suarez, Malte Ostendorff, Samuel Weinbach, Rafet Sifa, Stefan Kesselheim, and Nicolas Flores{-}Herr.
\newblock Tokenizer choice for {LLM} training: Negligible or crucial?
\newblock In Kevin Duh, Helena G{\'{o}}mez{-}Adorno, and Steven Bethard (eds.), \emph{Findings of the Association for Computational Linguistics: {NAACL} 2024, Mexico City, Mexico, June 16-21, 2024}, volume {NAACL} 2024 of \emph{Findings of {ACL}}, pp.\  3907--3924. Association for Computational Linguistics, 2024.
\newblock \doi{10.18653/V1/2024.FINDINGS-NAACL.247}.
\newblock URL \url{https://doi.org/10.18653/v1/2024.findings-naacl.247}.

\bibitem[Boizard et~al.(2025)Boizard, Haddad, Hudelot, and Colombo]{boizard2025crosstokenizerdistillationuniversallogit}
Nicolas Boizard, Kevin~El Haddad, Céline Hudelot, and Pierre Colombo.
\newblock Towards cross-tokenizer distillation: the universal logit distillation loss for llms, 2025.
\newblock URL \url{https://arxiv.org/abs/2402.12030}.

\bibitem[Chen et~al.(2026)Chen, Ruan, Dan, Wang, Yan, Wu, Zhang, Chen, Zhou, He, Qi, Li, Guo, Shi, and Zhang]{chen2026surveyinductivereasoninglarge}
Kedi Chen, Dezhao Ruan, Yuhao Dan, Yaoting Wang, Siyu Yan, Xuecheng Wu, Yinqi Zhang, Qin Chen, Jie Zhou, Liang He, Biqing Qi, Linyang Li, Qipeng Guo, Xiaoming Shi, and Wei Zhang.
\newblock A survey of inductive reasoning for large language models, 2026.
\newblock URL \url{https://arxiv.org/abs/2510.10182}.

\bibitem[Chen et~al.(2025)Chen, Liu, Meng, Chen, Xu, and Zhou]{chen2025enhancingcrosstokenizerknowledgedistillation}
Yijie Chen, Yijin Liu, Fandong Meng, Yufeng Chen, Jinan Xu, and Jie Zhou.
\newblock Enhancing cross-tokenizer knowledge distillation with contextual dynamical mapping, 2025.
\newblock URL \url{https://arxiv.org/abs/2502.11104}.

\bibitem[Clark et~al.(2019)Clark, Lee, Chang, Kwiatkowski, Collins, and Toutanova]{clark-etal-2019-boolq}
Christopher Clark, Kenton Lee, Ming-Wei Chang, Tom Kwiatkowski, Michael Collins, and Kristina Toutanova.
\newblock {B}ool{Q}: Exploring the surprising difficulty of natural yes/no questions.
\newblock In Jill Burstein, Christy Doran, and Thamar Solorio (eds.), \emph{Proceedings of the 2019 Conference of the North {A}merican Chapter of the Association for Computational Linguistics: Human Language Technologies, Volume 1 (Long and Short Papers)}, pp.\  2924--2936, Minneapolis, Minnesota, June 2019. Association for Computational Linguistics.
\newblock \doi{10.18653/v1/N19-1300}.
\newblock URL \url{https://aclanthology.org/N19-1300/}.

\bibitem[Clark et~al.(2018)Clark, Cowhey, Etzioni, Khot, Sabharwal, Schoenick, and Tafjord]{clark2018thinksolvedquestionanswering}
Peter Clark, Isaac Cowhey, Oren Etzioni, Tushar Khot, Ashish Sabharwal, Carissa Schoenick, and Oyvind Tafjord.
\newblock Think you have solved question answering? try arc, the ai2 reasoning challenge, 2018.
\newblock URL \url{https://arxiv.org/abs/1803.05457}.

\bibitem[Cobbe et~al.(2021)Cobbe, Kosaraju, Bavarian, Chen, Jun, Kaiser, Plappert, Tworek, Hilton, Nakano, Hesse, and Schulman]{cobbe2021trainingverifierssolvemath}
Karl Cobbe, Vineet Kosaraju, Mohammad Bavarian, Mark Chen, Heewoo Jun, Lukasz Kaiser, Matthias Plappert, Jerry Tworek, Jacob Hilton, Reiichiro Nakano, Christopher Hesse, and John Schulman.
\newblock Training verifiers to solve math word problems, 2021.
\newblock URL \url{https://arxiv.org/abs/2110.14168}.

\bibitem[Cui et~al.(2025)Cui, Zhu, Qin, Xie, Zhou, and Li]{cui2025multileveloptimaltransportuniversal}
Xiao Cui, Mo~Zhu, Yulei Qin, Liang Xie, Wengang Zhou, and Houqiang Li.
\newblock Multi-level optimal transport for universal cross-tokenizer knowledge distillation on language models, 2025.
\newblock URL \url{https://arxiv.org/abs/2412.14528}.

\bibitem[Dao et~al.(2026)Dao, Nguyen, Chi, Nguyen, Van, Diep, and Le]{dao2026sraspanrepresentationalignment}
Quoc~Phong Dao, Hoang~Son Nguyen, Pham~Khanh Chi, Tung Nguyen, Linh~Ngo Van, Nguyen Thi~Ngoc Diep, and Trung Le.
\newblock Sra: Span representation alignment for large language model distillation, 2026.
\newblock URL \url{https://arxiv.org/abs/2605.01205}.

\bibitem[Gandhi et~al.(2024)Gandhi, Lee, Grand, Liu, Cheng, Sharma, and Goodman]{gandhi2024streamsearchsoslearning}
Kanishk Gandhi, Denise Lee, Gabriel Grand, Muxin Liu, Winson Cheng, Archit Sharma, and Noah~D. Goodman.
\newblock Stream of search (sos): Learning to search in language, 2024.
\newblock URL \url{https://arxiv.org/abs/2404.03683}.

\bibitem[Goddard \& Neto(2025)Goddard and Neto]{goddard2025trainingfreetokenizertransplantationorthogonal}
Charles Goddard and Fernando~Fernandes Neto.
\newblock Training-free tokenizer transplantation via orthogonal matching pursuit, 2025.
\newblock URL \url{https://arxiv.org/abs/2506.06607}.

\bibitem[Gu et~al.(2024)Gu, Rozière, Leather, Solar-Lezama, Synnaeve, and Wang]{gu2024cruxevalbenchmarkcodereasoning}
Alex Gu, Baptiste Rozière, Hugh Leather, Armando Solar-Lezama, Gabriel Synnaeve, and Sida~I. Wang.
\newblock Cruxeval: A benchmark for code reasoning, understanding and execution, 2024.
\newblock URL \url{https://arxiv.org/abs/2401.03065}.

\bibitem[Gu et~al.(2026)Gu, Dong, Wei, and Huang]{gu2026minillmonpolicydistillationlarge}
Yuxian Gu, Li~Dong, Furu Wei, and Minlie Huang.
\newblock Minillm: On-policy distillation of large language models, 2026.
\newblock URL \url{https://arxiv.org/abs/2306.08543}.

\bibitem[Havrilla et~al.(2024)Havrilla, Dai, O'Mahony, Oostermeijer, Zisler, Albalak, Milo, Raparthy, Gandhi, Abbasi, Phung, Iyer, Mahan, Blagden, Gureja, Hamdy, Li, Paolini, Ammanamanchi, and Meyerson]{havrilla2024surveyingeffectsqualitydiversity}
Alex Havrilla, Andrew Dai, Laura O'Mahony, Koen Oostermeijer, Vera Zisler, Alon Albalak, Fabrizio Milo, Sharath~Chandra Raparthy, Kanishk Gandhi, Baber Abbasi, Duy Phung, Maia Iyer, Dakota Mahan, Chase Blagden, Srishti Gureja, Mohammed Hamdy, Wen-Ding Li, Giovanni Paolini, Pawan~Sasanka Ammanamanchi, and Elliot Meyerson.
\newblock Surveying the effects of quality, diversity, and complexity in synthetic data from large language models, 2024.
\newblock URL \url{https://arxiv.org/abs/2412.02980}.

\bibitem[He et~al.(2025)He, Liu, Liu, Yan, Wang, Cheng, Zhang, Zhang, Xu, Shen, Li, Zeng, Wei, Cheng, An, Liu, and Zhou]{he2025skyworkopenreasoner1}
Jujie He, Jiacai Liu, Chris~Yuhao Liu, Rui Yan, Chaojie Wang, Peng Cheng, Xiaoyu Zhang, Fuxiang Zhang, Jiacheng Xu, Wei Shen, Siyuan Li, Liang Zeng, Tianwen Wei, Cheng Cheng, Bo~An, Yang Liu, and Yahui Zhou.
\newblock Skywork open reasoner 1 technical report, 2025.
\newblock URL \url{https://arxiv.org/abs/2505.22312}.

\bibitem[Hendrycks et~al.(2021)Hendrycks, Burns, Basart, Zou, Mazeika, Song, and Steinhardt]{hendrycks2021measuringmassivemultitasklanguage}
Dan Hendrycks, Collin Burns, Steven Basart, Andy Zou, Mantas Mazeika, Dawn Song, and Jacob Steinhardt.
\newblock Measuring massive multitask language understanding, 2021.
\newblock URL \url{https://arxiv.org/abs/2009.03300}.

\bibitem[Hinton et~al.(2015)Hinton, Vinyals, and Dean]{hinton2015distillingknowledgeneuralnetwork}
Geoffrey Hinton, Oriol Vinyals, and Jeff Dean.
\newblock Distilling the knowledge in a neural network, 2015.
\newblock URL \url{https://arxiv.org/abs/1503.02531}.

\bibitem[Jain et~al.(2024)Jain, Han, Gu, Li, Yan, Zhang, Wang, Solar-Lezama, Sen, and Stoica]{jain2024livecodebenchholisticcontaminationfree}
Naman Jain, King Han, Alex Gu, Wen-Ding Li, Fanjia Yan, Tianjun Zhang, Sida Wang, Armando Solar-Lezama, Koushik Sen, and Ion Stoica.
\newblock Livecodebench: Holistic and contamination free evaluation of large language models for code, 2024.
\newblock URL \url{https://arxiv.org/abs/2403.07974}.

\bibitem[Jia et~al.(2025)Jia, Gao, Xue, Wang, Cai, Chen, Zhao, Jiang, and Gai]{jia2025principlesapplicationscomprehensivesurvey}
Jian Jia, Jingtong Gao, Ben Xue, Junhao Wang, Qingpeng Cai, Quan Chen, Xiangyu Zhao, Peng Jiang, and Kun Gai.
\newblock From principles to applications: A comprehensive survey of discrete tokenizers in generation, comprehension, recommendation, and information retrieval, 2025.
\newblock URL \url{https://arxiv.org/abs/2502.12448}.

\bibitem[Ko et~al.(2024)Ko, Kim, Chen, and Yun]{ko2024distillmstreamlineddistillationlarge}
Jongwoo Ko, Sungnyun Kim, Tianyi Chen, and Se-Young Yun.
\newblock Distillm: Towards streamlined distillation for large language models, 2024.
\newblock URL \url{https://arxiv.org/abs/2402.03898}.

\bibitem[Kwon et~al.(2023)Kwon, Li, Zhuang, Sheng, Zheng, Yu, Gonzalez, Zhang, and Stoica]{kwon2023efficient}
Woosuk Kwon, Zhuohan Li, Siyuan Zhuang, Ying Sheng, Lianmin Zheng, Cody~Hao Yu, Joseph~E. Gonzalez, Hao Zhang, and Ion Stoica.
\newblock Efficient memory management for large language model serving with pagedattention.
\newblock In \emph{Proceedings of the ACM SIGOPS 29th Symposium on Operating Systems Principles}, 2023.

\bibitem[Le et~al.(2025)Le, Vu, Hai, Diep, Van, Le, and Nguyen]{le2025cot2aligncrosschainthoughtdistillation}
Anh~Duc Le, Tu~Vu, Nam~Le Hai, Nguyen Thi~Ngoc Diep, Linh~Ngo Van, Trung Le, and Thien~Huu Nguyen.
\newblock Cot2align: Cross-chain of thought distillation via optimal transport alignment for language models with different tokenizers, 2025.
\newblock URL \url{https://arxiv.org/abs/2502.16806}.

\bibitem[Li et~al.(2025)Li, Zhang, and Zong]{li-etal-2025-tokalign}
Chong Li, Jiajun Zhang, and Chengqing Zong.
\newblock {T}ok{A}lign: Efficient vocabulary adaptation via token alignment.
\newblock In Wanxiang Che, Joyce Nabende, Ekaterina Shutova, and Mohammad~Taher Pilehvar (eds.), \emph{Proceedings of the 63rd Annual Meeting of the Association for Computational Linguistics (Volume 1: Long Papers)}, pp.\  4109--4126, Vienna, Austria, July 2025. Association for Computational Linguistics.
\newblock ISBN 979-8-89176-251-0.
\newblock \doi{10.18653/v1/2025.acl-long.207}.
\newblock URL \url{https://aclanthology.org/2025.acl-long.207/}.

\bibitem[Li et~al.(2023)Li, Fu, Zhang, Huang, Sun, Lyu, Liu, Jin, and Li]{li2023tacotopicsalgorithmiccode}
Rongao Li, Jie Fu, Bo-Wen Zhang, Tao Huang, Zhihong Sun, Chen Lyu, Guang Liu, Zhi Jin, and Ge~Li.
\newblock Taco: Topics in algorithmic code generation dataset, 2023.
\newblock URL \url{https://arxiv.org/abs/2312.14852}.

\bibitem[Lightman et~al.(2023)Lightman, Kosaraju, Burda, Edwards, Baker, Lee, Leike, Schulman, Sutskever, and Cobbe]{lightman2023letsverifystepstep}
Hunter Lightman, Vineet Kosaraju, Yura Burda, Harri Edwards, Bowen Baker, Teddy Lee, Jan Leike, John Schulman, Ilya Sutskever, and Karl Cobbe.
\newblock Let's verify step by step, 2023.
\newblock URL \url{https://arxiv.org/abs/2305.20050}.

\bibitem[Lin et~al.(2026)Lin, Chen, and Zhang]{lin2026renioreweightingnegativetrajectory}
Chen Lin, Kedi Chen, and Wei Zhang.
\newblock Renio: Reweighting negative trajectory importance for llm on-policy distillation, 2026.
\newblock URL \url{https://arxiv.org/abs/2606.23104}.

\bibitem[Liu et~al.(2025)Liu, Hayase, Hofmann, Oh, Smith, and Choi]{liu2025superbpespacetravellanguage}
Alisa Liu, Jonathan Hayase, Valentin Hofmann, Sewoong Oh, Noah~A. Smith, and Yejin Choi.
\newblock Superbpe: Space travel for language models, 2025.
\newblock URL \url{https://arxiv.org/abs/2503.13423}.

\bibitem[Liu et~al.(2024)Liu, Wang, Qing, Kuang, Kang, Sun, and Wu]{DBLP:conf/emnlp/LiuWQKKS024}
Chengyuan Liu, Shihang Wang, Lizhi Qing, Kun Kuang, Yangyang Kang, Changlong Sun, and Fei Wu.
\newblock Gold panning in vocabulary: An adaptive method for vocabulary expansion of domain-specific llms.
\newblock In Yaser Al{-}Onaizan, Mohit Bansal, and Yun{-}Nung Chen (eds.), \emph{Proceedings of the 2024 Conference on Empirical Methods in Natural Language Processing, {EMNLP} 2024, Miami, FL, USA, November 12-16, 2024}, pp.\  7442--7459. Association for Computational Linguistics, 2024.
\newblock \doi{10.18653/V1/2024.EMNLP-MAIN.424}.
\newblock URL \url{https://doi.org/10.18653/v1/2024.emnlp-main.424}.

\bibitem[Liu et~al.(2023)Liu, Xia, Wang, and Zhang]{liu2023codegeneratedchatgptreally}
Jiawei Liu, Chunqiu~Steven Xia, Yuyao Wang, and Lingming Zhang.
\newblock Is your code generated by chatgpt really correct? rigorous evaluation of large language models for code generation, 2023.
\newblock URL \url{https://arxiv.org/abs/2305.01210}.

\bibitem[Minixhofer et~al.(2025)Minixhofer, Vulić, and Ponti]{minixhofer2025universalcrosstokenizerdistillationapproximate}
Benjamin Minixhofer, Ivan Vulić, and Edoardo~Maria Ponti.
\newblock Universal cross-tokenizer distillation via approximate likelihood matching, 2025.
\newblock URL \url{https://arxiv.org/abs/2503.20083}.

\bibitem[Nguyen et~al.(2026)Nguyen, Dat, Nguyen, Van, Le, and Nguyen]{nguyen2026ctpdcrosstokenizerpreference}
Truong Nguyen, Phi~Van Dat, Ngan Nguyen, Linh~Ngo Van, Trung Le, and Thanh~Hong Nguyen.
\newblock Ctpd: Cross tokenizer preference distillation, 2026.
\newblock URL \url{https://arxiv.org/abs/2601.11865}.

\bibitem[NVIDIA et~al.(2025)NVIDIA, :, Basant, Khairnar, Paithankar, Khattar, Renduchintala, Malte, Bercovich, Hazare, Rico, Ficek, Kondratenko, Shaposhnikov, Bukharin, Taghibakhshi, Barton, Mahabaleshwarkar, Shen, Tao, Guan, Shors, Mandarwal, Mehta, Venkatesan, Sharabiani, Aithal, Poojary, Dattagupta, Buddharaju, Zhu, Simkin, Kartal, Rouhani, Chen, Ginsburg, Norick, Yu, Catanzaro, Wang, Truong, Mungekar, Patel, Alexiuk, Munley, Parisien, Su, Afrimi, Korzekwa, Rohrer, Gitman, Mosallanezhad, Narayanan, Rekesh, Yared, Pykhtar, Ahn, Riach, Long, Ning, Chung, Galinkin, Bakhturina, Prasad, Shen, Qian, Elisha, Sharma, Ross, Ngo, Sahota, Wang, Shin, Huang, Cunningham, Gitman, Moshkov, Jung, Kautz, Scowcroft, Casper, Zhang, Zeng, Zhang, Xue, Huang, Conway, Kamalu, Cohen, Jennings, Vialard, Yi, Parmar, Briski, Cheung, Luna, Wyss, Santhanam, Kong, Pawelec, Anik, Li, Ahmadian, McAfee, Sleiman, Derczynski, Vega, de~Melo, Sreedhar, Chochowski, Cai, Kliegl, Stepniewska-Dziubinska, Novikov, Samadi, Price, Boubdir, Boone,
  Evans, Bien, Zawalski, Martinez, Chrzanowski, Shoeybi, Patwary, Dhameja, Assaf, Habibi, Bhatia, Pope, Tajbakhsh, Juluru, Rybakov, Hrinchuk, Kuchaiev, Olabiyi, Ribalta, Subramanian, Chadha, Molchanov, Dykas, Jin, Bialecki, Januszewski, Thalasta, Gaikwad, Varshney, Gundecha, Tredak, Mahabadi, Patel, El-Yaniv, Rajan, Cheruvu, Shahbazyan, Borkar, Gala, Waleffe, Zhang, Hewett, Prenger, Jain, Kriman, Satheesh, Kaji, Yurick, Muralidharan, Narenthiran, Bak, Sameni, Han, Ramasamy, Ghosh, Sreenivas, Thomas, Diao, Gopal, Prabhumoye, Toshniwal, Ding, Singh, Jain, Majumdar, Singhal, Alborghetti, Akter, Kong, Moon, Hliwiak, Asida, Wang, Konuk, Vashishth, Poon, Karpas, Noroozi, Srinivasan, Korthikanti, Fugro, Kalluru, Kurin, Lavrukhin, Ahmad, Du, Byeon, Lu, Dong, Karnati, Choi, Zhang, Lin, Fu, Suhara, Dong, Li, Zhu, and Chen]{nvidia2025nvidianemotronnano2}
NVIDIA, :, Aarti Basant, Abhijit Khairnar, Abhijit Paithankar, Abhinav Khattar, Adithya Renduchintala, Aditya Malte, Akhiad Bercovich, Akshay Hazare, Alejandra Rico, Aleksander Ficek, Alex Kondratenko, Alex Shaposhnikov, Alexander Bukharin, Ali Taghibakhshi, Amelia Barton, Ameya~Sunil Mahabaleshwarkar, Amy Shen, Andrew Tao, Ann Guan, Anna Shors, Anubhav Mandarwal, Arham Mehta, Arun Venkatesan, Ashton Sharabiani, Ashwath Aithal, Ashwin Poojary, Ayush Dattagupta, Balaram Buddharaju, Banghua Zhu, Barnaby Simkin, Bilal Kartal, Bita~Darvish Rouhani, Bobby Chen, Boris Ginsburg, Brandon Norick, Brian Yu, Bryan Catanzaro, Charles Wang, Charlie Truong, Chetan Mungekar, Chintan Patel, Chris Alexiuk, Christian Munley, Christopher Parisien, Dan Su, Daniel Afrimi, Daniel Korzekwa, Daniel Rohrer, Daria Gitman, David Mosallanezhad, Deepak Narayanan, Dima Rekesh, Dina Yared, Dmytro Pykhtar, Dong Ahn, Duncan Riach, Eileen Long, Elliott Ning, Eric Chung, Erick Galinkin, Evelina Bakhturina, Gargi Prasad, Gerald Shen, Haifeng
  Qian, Haim Elisha, Harsh Sharma, Hayley Ross, Helen Ngo, Herman Sahota, Hexin Wang, Hoo~Chang Shin, Hua Huang, Iain Cunningham, Igor Gitman, Ivan Moshkov, Jaehun Jung, Jan Kautz, Jane~Polak Scowcroft, Jared Casper, Jian Zhang, Jiaqi Zeng, Jimmy Zhang, Jinze Xue, Jocelyn Huang, Joey Conway, John Kamalu, Jonathan Cohen, Joseph Jennings, Julien~Veron Vialard, Junkeun Yi, Jupinder Parmar, Kari Briski, Katherine Cheung, Katherine Luna, Keith Wyss, Keshav Santhanam, Kezhi Kong, Krzysztof Pawelec, Kumar Anik, Kunlun Li, Kushan Ahmadian, Lawrence McAfee, Laya Sleiman, Leon Derczynski, Luis Vega, Maer~Rodrigues de~Melo, Makesh~Narsimhan Sreedhar, Marcin Chochowski, Mark Cai, Markus Kliegl, Marta Stepniewska-Dziubinska, Matvei Novikov, Mehrzad Samadi, Meredith Price, Meriem Boubdir, Michael Boone, Michael Evans, Michal Bien, Michal Zawalski, Miguel Martinez, Mike Chrzanowski, Mohammad Shoeybi, Mostofa Patwary, Namit Dhameja, Nave Assaf, Negar Habibi, Nidhi Bhatia, Nikki Pope, Nima Tajbakhsh, Nirmal~Kumar Juluru, Oleg
  Rybakov, Oleksii Hrinchuk, Oleksii Kuchaiev, Oluwatobi Olabiyi, Pablo Ribalta, Padmavathy Subramanian, Parth Chadha, Pavlo Molchanov, Peter Dykas, Peter Jin, Piotr Bialecki, Piotr Januszewski, Pradeep Thalasta, Prashant Gaikwad, Prasoon Varshney, Pritam Gundecha, Przemek Tredak, Rabeeh~Karimi Mahabadi, Rajen Patel, Ran El-Yaniv, Ranjit Rajan, Ria Cheruvu, Rima Shahbazyan, Ritika Borkar, Ritu Gala, Roger Waleffe, Ruoxi Zhang, Russell~J. Hewett, Ryan Prenger, Sahil Jain, Samuel Kriman, Sanjeev Satheesh, Saori Kaji, Sarah Yurick, Saurav Muralidharan, Sean Narenthiran, Seonmyeong Bak, Sepehr Sameni, Seungju Han, Shanmugam Ramasamy, Shaona Ghosh, Sharath~Turuvekere Sreenivas, Shelby Thomas, Shizhe Diao, Shreya Gopal, Shrimai Prabhumoye, Shubham Toshniwal, Shuoyang Ding, Siddharth Singh, Siddhartha Jain, Somshubra Majumdar, Soumye Singhal, Stefania Alborghetti, Syeda~Nahida Akter, Terry Kong, Tim Moon, Tomasz Hliwiak, Tomer Asida, Tony Wang, Tugrul Konuk, Twinkle Vashishth, Tyler Poon, Udi Karpas, Vahid Noroozi,
  Venkat Srinivasan, Vijay Korthikanti, Vikram Fugro, Vineeth Kalluru, Vitaly Kurin, Vitaly Lavrukhin, Wasi~Uddin Ahmad, Wei Du, Wonmin Byeon, Ximing Lu, Xin Dong, Yashaswi Karnati, Yejin Choi, Yian Zhang, Ying Lin, Yonggan Fu, Yoshi Suhara, Zhen Dong, Zhiyu Li, Zhongbo Zhu, and Zijia Chen.
\newblock Nvidia nemotron nano 2: An accurate and efficient hybrid mamba-transformer reasoning model, 2025.
\newblock URL \url{https://arxiv.org/abs/2508.14444}.

\bibitem[OLMo et~al.(2025)OLMo, Walsh, Soldaini, Groeneveld, Lo, Arora, Bhagia, Gu, Huang, Jordan, Lambert, Schwenk, Tafjord, Anderson, Atkinson, Brahman, Clark, Dasigi, Dziri, Ettinger, Guerquin, Heineman, Ivison, Koh, Liu, Malik, Merrill, Miranda, Morrison, Murray, Nam, Poznanski, Pyatkin, Rangapur, Schmitz, Skjonsberg, Wadden, Wilhelm, Wilson, Zettlemoyer, Farhadi, Smith, and Hajishirzi]{olmo20252olmo2furious}
Team OLMo, Pete Walsh, Luca Soldaini, Dirk Groeneveld, Kyle Lo, Shane Arora, Akshita Bhagia, Yuling Gu, Shengyi Huang, Matt Jordan, Nathan Lambert, Dustin Schwenk, Oyvind Tafjord, Taira Anderson, David Atkinson, Faeze Brahman, Christopher Clark, Pradeep Dasigi, Nouha Dziri, Allyson Ettinger, Michal Guerquin, David Heineman, Hamish Ivison, Pang~Wei Koh, Jiacheng Liu, Saumya Malik, William Merrill, Lester James~V. Miranda, Jacob Morrison, Tyler Murray, Crystal Nam, Jake Poznanski, Valentina Pyatkin, Aman Rangapur, Michael Schmitz, Sam Skjonsberg, David Wadden, Christopher Wilhelm, Michael Wilson, Luke Zettlemoyer, Ali Farhadi, Noah~A. Smith, and Hannaneh Hajishirzi.
\newblock 2 olmo 2 furious, 2025.
\newblock URL \url{https://arxiv.org/abs/2501.00656}.

\bibitem[Patiño et~al.(2025)Patiño, Rasul, Gallouédec, Burtenshaw, Paniego, Srivastav, Frere, Beeching, Tunstall, von Werra, and Wolf]{patiño2025_unlocking_on_policy_distillation_for_any_model_family}
Carlos~Miguel Patiño, Kashif Rasul, Quentin Gallouédec, Ben Burtenshaw, Sergio Paniego, Vaibhav Srivastav, Thibaud Frere, Ed~Beeching, Lewis Tunstall, Leandro von Werra, and Thomas Wolf.
\newblock Unlocking on-policy distillation for any model family.
\newblock \url{https://huggingface.co/spaces/HuggingFaceH4/on-policy-distillation}, 2025.

\bibitem[Phan et~al.(2026)Phan, Khisti, and Ullrich]{phan2026crosstokenizerlikelihoodscoringalgorithms}
Buu Phan, Ashish Khisti, and Karen Ullrich.
\newblock Cross-tokenizer likelihood scoring algorithms for language model distillation, 2026.
\newblock URL \url{https://arxiv.org/abs/2512.14954}.

\bibitem[Sakaguchi et~al.(2019)Sakaguchi, Bras, Bhagavatula, and Choi]{sakaguchi2019winograndeadversarialwinogradschema}
Keisuke Sakaguchi, Ronan~Le Bras, Chandra Bhagavatula, and Yejin Choi.
\newblock Winogrande: An adversarial winograd schema challenge at scale, 2019.
\newblock URL \url{https://arxiv.org/abs/1907.10641}.

\bibitem[Shao et~al.(2024{\natexlab{a}})Shao, Basit, Karri, and Shafique]{Shao_2024}
Minghao Shao, Abdul Basit, Ramesh Karri, and Muhammad Shafique.
\newblock Survey of different large language model architectures: Trends, benchmarks, and challenges.
\newblock \emph{IEEE Access}, 12:\penalty0 188664–188706, 2024{\natexlab{a}}.
\newblock ISSN 2169-3536.
\newblock \doi{10.1109/access.2024.3482107}.
\newblock URL \url{http://dx.doi.org/10.1109/ACCESS.2024.3482107}.

\bibitem[Shao et~al.(2024{\natexlab{b}})Shao, Wang, Zhu, Xu, Song, Bi, Zhang, Zhang, Li, Wu, and Guo]{shao2024deepseekmathpushinglimitsmathematical}
Zhihong Shao, Peiyi Wang, Qihao Zhu, Runxin Xu, Junxiao Song, Xiao Bi, Haowei Zhang, Mingchuan Zhang, Y.~K. Li, Y.~Wu, and Daya Guo.
\newblock Deepseekmath: Pushing the limits of mathematical reasoning in open language models, 2024{\natexlab{b}}.
\newblock URL \url{https://arxiv.org/abs/2402.03300}.

\bibitem[Shin et~al.(2025)Shin, Ji, Liu, and Gong]{DBLP:conf/icml/Shin0LG25}
Haebin Shin, Lei Ji, Xiao Liu, and Yeyun Gong.
\newblock Overcoming vocabulary mismatch: Vocabulary-agnostic teacher guided language modeling.
\newblock In Aarti Singh, Maryam Fazel, Daniel Hsu, Simon Lacoste{-}Julien, Felix Berkenkamp, Tegan Maharaj, Kiri Wagstaff, and Jerry Zhu (eds.), \emph{Forty-second International Conference on Machine Learning, {ICML} 2025, Vancouver, BC, Canada, July 13-19, 2025}, volume 267 of \emph{Proceedings of Machine Learning Research}. {PMLR} / OpenReview.net, 2025.
\newblock URL \url{https://proceedings.mlr.press/v267/shin25a.html}.

\bibitem[Singh et~al.(2026)Singh, Wu, Cioba, Bernacchia, and Buffelli]{singh2026crosstokenizerllmdistillationbytelevel}
Avyav~Kumar Singh, Yen-Chen Wu, Alexandru Cioba, Alberto Bernacchia, and Davide Buffelli.
\newblock Cross-tokenizer llm distillation through a byte-level interface, 2026.
\newblock URL \url{https://arxiv.org/abs/2604.07466}.

\bibitem[Song \& Zheng(2026)Song and Zheng]{song2026surveyonpolicydistillationlarge}
Mingyang Song and Mao Zheng.
\newblock A survey of on-policy distillation for large language models, 2026.
\newblock URL \url{https://arxiv.org/abs/2604.00626}.

\bibitem[Sreenivas et~al.(2026)Sreenivas, Hanasoge, Yang, Taghibakhshi, Muralidharan, Aithal, and Molchanov]{sreenivas2026xtokenprojectionguidedcrosstokenizerknowledge}
Sharath~Turuvekere Sreenivas, Adithyakrishna~Venkatesh Hanasoge, Mingyu Yang, Ali Taghibakhshi, Saurav Muralidharan, Ashwath Aithal, and Pavlo Molchanov.
\newblock X-token: Projection-guided cross-tokenizer knowledge distillation, 2026.
\newblock URL \url{https://arxiv.org/abs/2605.21699}.

\bibitem[Sun et~al.(2026)Sun, Zheng, Song, Zhong, Cheng, Feng, Liu, Fang, and Wang]{sun2026simctrecoveringlostsupervision}
Jie Sun, Mao Zheng, Mingyang Song, Qiyong Zhong, Yilin Cheng, Bichuan Feng, Pengfei Liu, Junfeng Fang, and Xiang Wang.
\newblock Simct: Recovering lost supervision for cross-tokenizer on-policy distillation, 2026.
\newblock URL \url{https://arxiv.org/abs/2605.07711}.

\bibitem[Team(2024)]{DBLP:journals/corr/abs-2407-21783}
Llama Team.
\newblock The llama 3 herd of models.
\newblock \emph{CoRR}, abs/2407.21783, 2024.
\newblock \doi{10.48550/ARXIV.2407.21783}.
\newblock URL \url{https://doi.org/10.48550/arXiv.2407.21783}.

\bibitem[Tirumala et~al.(2023)Tirumala, Simig, Aghajanyan, and Morcos]{DBLP:conf/nips/TirumalaSAM23}
Kushal Tirumala, Daniel Simig, Armen Aghajanyan, and Ari Morcos.
\newblock {D4:} improving {LLM} pretraining via document de-duplication and diversification.
\newblock In Alice Oh, Tristan Naumann, Amir Globerson, Kate Saenko, Moritz Hardt, and Sergey Levine (eds.), \emph{Advances in Neural Information Processing Systems 36: Annual Conference on Neural Information Processing Systems 2023, NeurIPS 2023, New Orleans, LA, USA, December 10 - 16, 2023}, 2023.
\newblock URL \url{http://papers.nips.cc/paper\_files/paper/2023/hash/a8f8cbd7f7a5fb2c837e578c75e5b615-Abstract-Datasets\_and\_Benchmarks.html}.

\bibitem[Vu et~al.(2026)Vu, Chi, Van, Van, Dinh, and Le]{vu2026dwakddualspaceweightingtimewarped}
Duc~Trung Vu, Pham~Khanh Chi, Dat~Phi Van, Linh~Ngo Van, Sang Dinh, and Trung Le.
\newblock Dwa-kd: Dual-space weighting and time-warped alignment for cross-tokenizer knowledge distillation, 2026.
\newblock URL \url{https://arxiv.org/abs/2602.21669}.

\bibitem[Vuong et~al.(2026)Vuong, Le, Tran, Van, and Le]{DBLP:conf/aaai/VuongLTVL26}
Hoang~Tran Vuong, Tue Le, Quyen Tran, Linh~Ngo Van, and Trung Le.
\newblock {MCW-KD:} multi-cost wasserstein knowledge distillation for large language models.
\newblock In Sven Koenig, Chad Jenkins, and Matthew~E. Taylor (eds.), \emph{Fortieth {AAAI} Conference on Artificial Intelligence, Thirty-Eighth Conference on Innovative Applications of Artificial Intelligence, Sixteenth Symposium on Educational Advances in Artificial Intelligence, {AAAI} 2026, Singapore, January 20-27, 2026}, pp.\  33332--33340. {AAAI} Press, 2026.
\newblock \doi{10.1609/AAAI.V40I39.40619}.
\newblock URL \url{https://doi.org/10.1609/aaai.v40i39.40619}.

\bibitem[Wang et~al.(2026{\natexlab{a}})Wang, Yuan, Zhong, Zhang, Xiao, Sun, and Qi]{wang2026crosstokenizeronpolicydistillationbyteprefix}
Hao Wang, Kun Yuan, Wenlin Zhong, Minglei Zhang, Han Xiao, Ming Sun, and Honggang Qi.
\newblock Cross-tokenizer on-policy distillation via byte-prefix marginalization, 2026{\natexlab{a}}.
\newblock URL \url{https://arxiv.org/abs/2607.22334}.

\bibitem[Wang et~al.(2026{\natexlab{b}})Wang, Jing, Sun, Wang, Liao, Rutkowski, and Tao]{DBLP:conf/aaai/WangJSWLRT26}
Huazheng Wang, Yongcheng Jing, Haifeng Sun, Jingyu Wang, Jianxin Liao, Leszek Rutkowski, and Dacheng Tao.
\newblock Bridging the tokenizer gap: Semantics and distribution-aware knowledge transfer for unbiased cross-tokenizer distillation.
\newblock In Sven Koenig, Chad Jenkins, and Matthew~E. Taylor (eds.), \emph{Fortieth {AAAI} Conference on Artificial Intelligence, Thirty-Eighth Conference on Innovative Applications of Artificial Intelligence, Sixteenth Symposium on Educational Advances in Artificial Intelligence, {AAAI} 2026, Singapore, January 20-27, 2026}, pp.\  33494--33502. {AAAI} Press, 2026{\natexlab{b}}.
\newblock \doi{10.1609/AAAI.V40I39.40637}.
\newblock URL \url{https://doi.org/10.1609/aaai.v40i39.40637}.

\bibitem[Wang et~al.(2025)Wang, Chen, Wang, U, Li, and Guo]{wang2025largelanguagemodelenhanced}
Xin Wang, Zirui Chen, Haofen Wang, Leong~Hou U, Zhao Li, and Wenbin Guo.
\newblock Large language model enhanced knowledge representation learning: A survey, 2025.
\newblock URL \url{https://arxiv.org/abs/2407.00936}.

\bibitem[Xia et~al.(2025)Xia, Shen, Wang, Liu, Sun, Wu, Hu, and Xu]{xia2025leetcodedatasettemporaldatasetrobust}
Yunhui Xia, Wei Shen, Yan Wang, Jason~Klein Liu, Huifeng Sun, Siyue Wu, Jian Hu, and Xiaolong Xu.
\newblock Leetcodedataset: A temporal dataset for robust evaluation and efficient training of code llms, 2025.
\newblock URL \url{https://arxiv.org/abs/2504.14655}.

\bibitem[Xu et~al.(2025)Xu, Hao, Zong, Wang, Zhang, Wang, Lan, Gong, Ouyang, Meng, Shao, Yan, Yang, Song, Ren, Hu, Li, Feng, Gao, and Li]{xu2025largereasoningmodelssurvey}
Fengli Xu, Qianyue Hao, Zefang Zong, Jingwei Wang, Yunke Zhang, Jingyi Wang, Xiaochong Lan, Jiahui Gong, Tianjian Ouyang, Fanjin Meng, Chenyang Shao, Yuwei Yan, Qinglong Yang, Yiwen Song, Sijian Ren, Xinyuan Hu, Yu~Li, Jie Feng, Chen Gao, and Yong Li.
\newblock Towards large reasoning models: A survey of reinforced reasoning with large language models, 2025.
\newblock URL \url{https://arxiv.org/abs/2501.09686}.

\bibitem[Yang et~al.(2024)Yang, Zhang, Hui, Gao, Yu, Li, Liu, Tu, Zhou, Lin, Lu, Xue, Lin, Liu, Ren, and Zhang]{yang2024qwen25mathtechnicalreportmathematical}
An~Yang, Beichen Zhang, Binyuan Hui, Bofei Gao, Bowen Yu, Chengpeng Li, Dayiheng Liu, Jianhong Tu, Jingren Zhou, Junyang Lin, Keming Lu, Mingfeng Xue, Runji Lin, Tianyu Liu, Xingzhang Ren, and Zhenru Zhang.
\newblock Qwen2.5-math technical report: Toward mathematical expert model via self-improvement.
\newblock \emph{arXiv preprint arXiv:2409.12122}, 2024.

\bibitem[Yang et~al.(2025)Yang, Li, Yang, Zhang, Hui, Zheng, Yu, Gao, Huang, Lv, Zheng, Liu, Zhou, Huang, Hu, Ge, Wei, Lin, Tang, Yang, Tu, Zhang, Yang, Yang, Zhou, Zhou, Lin, Dang, Bao, Yang, Yu, Deng, Li, Xue, Li, Zhang, Wang, Zhu, Men, Gao, Liu, Luo, Li, Tang, Yin, Ren, Wang, Zhang, Ren, Fan, Su, Zhang, Zhang, Wan, Liu, Wang, Cui, Zhang, Zhou, and Qiu]{yang2025qwen3technicalreport}
An~Yang, Anfeng Li, Baosong Yang, Beichen Zhang, Binyuan Hui, Bo~Zheng, Bowen Yu, Chang Gao, Chengen Huang, Chenxu Lv, Chujie Zheng, Dayiheng Liu, Fan Zhou, Fei Huang, Feng Hu, Hao Ge, Haoran Wei, Huan Lin, Jialong Tang, Jian Yang, Jianhong Tu, Jianwei Zhang, Jianxin Yang, Jiaxi Yang, Jing Zhou, Jingren Zhou, Junyang Lin, Kai Dang, Keqin Bao, Kexin Yang, Le~Yu, Lianghao Deng, Mei Li, Mingfeng Xue, Mingze Li, Pei Zhang, Peng Wang, Qin Zhu, Rui Men, Ruize Gao, Shixuan Liu, Shuang Luo, Tianhao Li, Tianyi Tang, Wenbiao Yin, Xingzhang Ren, Xinyu Wang, Xinyu Zhang, Xuancheng Ren, Yang Fan, Yang Su, Yichang Zhang, Yinger Zhang, Yu~Wan, Yuqiong Liu, Zekun Wang, Zeyu Cui, Zhenru Zhang, Zhipeng Zhou, and Zihan Qiu.
\newblock Qwen3 technical report, 2025.
\newblock URL \url{https://arxiv.org/abs/2505.09388}.

\bibitem[Yang et~al.(2026)Yang, Zhu, Song, Wang, Xia, Zheng, Ma, Chen, Wang, Zhao, and Chen]{yang2026oprdonpolicyrepresentationdistillation}
Shenzhi Yang, Guangcheng Zhu, Bowen Song, Haobo Wang, Mingxuan Xia, Xing Zheng, Yingfan Ma, Zhongqi Chen, Weiqiang Wang, Junbo Zhao, and Gang Chen.
\newblock Oprd: On-policy representation distillation, 2026.
\newblock URL \url{https://arxiv.org/abs/2606.06021}.

\bibitem[Yu et~al.(2025)Yu, Zhang, Zhu, Yuan, Zuo, Yue, Dai, Fan, Liu, Liu, Liu, Lin, Lin, Ma, Sheng, Tong, Zhang, Zhang, Zhang, Zhu, Zhu, Chen, Chen, Wang, Yu, Song, Wei, Zhou, Liu, Ma, Zhang, Yan, Qiao, Wu, and Wang]{yu2025dapoopensourcellmreinforcement}
Qiying Yu, Zheng Zhang, Ruofei Zhu, Yufeng Yuan, Xiaochen Zuo, Yu~Yue, Weinan Dai, Tiantian Fan, Gaohong Liu, Lingjun Liu, Xin Liu, Haibin Lin, Zhiqi Lin, Bole Ma, Guangming Sheng, Yuxuan Tong, Chi Zhang, Mofan Zhang, Wang Zhang, Hang Zhu, Jinhua Zhu, Jiaze Chen, Jiangjie Chen, Chengyi Wang, Hongli Yu, Yuxuan Song, Xiangpeng Wei, Hao Zhou, Jingjing Liu, Wei-Ying Ma, Ya-Qin Zhang, Lin Yan, Mu~Qiao, Yonghui Wu, and Mingxuan Wang.
\newblock Dapo: An open-source llm reinforcement learning system at scale, 2025.
\newblock URL \url{https://arxiv.org/abs/2503.14476}.

\bibitem[Zhang(2026)]{zhang2026formuladrivensurveyresearchagenda}
Bowen Zhang.
\newblock A formula-driven survey and research agenda for on-policy distillation, 2026.
\newblock URL \url{https://arxiv.org/abs/2606.22793}.

\bibitem[Zhang et~al.(2025{\natexlab{a}})Zhang, Zuo, He, Sun, Liu, Jiang, Fan, Tian, Jia, Li, Fu, Lv, Zhang, Zeng, Qu, Li, Wang, Wang, Long, Liu, Xu, Ma, Zhu, Hua, Liu, Li, Chen, Qu, Li, Chen, Yuan, Gao, Li, Ma, Cui, Liu, Qi, Ding, and Zhou]{zhang2025surveyreinforcementlearninglarge}
Kaiyan Zhang, Yuxin Zuo, Bingxiang He, Youbang Sun, Runze Liu, Che Jiang, Yuchen Fan, Kai Tian, Guoli Jia, Pengfei Li, Yu~Fu, Xingtai Lv, Yuchen Zhang, Sihang Zeng, Shang Qu, Haozhan Li, Shijie Wang, Yuru Wang, Xinwei Long, Fangfu Liu, Xiang Xu, Jiaze Ma, Xuekai Zhu, Ermo Hua, Yihao Liu, Zonglin Li, Huayu Chen, Xiaoye Qu, Yafu Li, Weize Chen, Zhenzhao Yuan, Junqi Gao, Dong Li, Zhiyuan Ma, Ganqu Cui, Zhiyuan Liu, Biqing Qi, Ning Ding, and Bowen Zhou.
\newblock A survey of reinforcement learning for large reasoning models, 2025{\natexlab{a}}.
\newblock URL \url{https://arxiv.org/abs/2509.08827}.

\bibitem[Zhang et~al.(2025{\natexlab{b}})Zhang, Dong, Li, Zhang, Sun, Wang, Li, Hu, Zhang, Wu, and Wang]{zhang2025instructiontuninglargelanguage}
Shengyu Zhang, Linfeng Dong, Xiaoya Li, Sen Zhang, Xiaofei Sun, Shuhe Wang, Jiwei Li, Runyi Hu, Tianwei Zhang, Fei Wu, and Guoyin Wang.
\newblock Instruction tuning for large language models: A survey, 2025{\natexlab{b}}.
\newblock URL \url{https://arxiv.org/abs/2308.10792}.

\bibitem[Zhang et~al.(2024)Zhang, Zhang, Sun, Chen, and Xu]{zhang2024dualspaceknowledgedistillationlarge}
Songming Zhang, Xue Zhang, Zengkui Sun, Yufeng Chen, and Jinan Xu.
\newblock Dual-space knowledge distillation for large language models, 2024.
\newblock URL \url{https://arxiv.org/abs/2406.17328}.

\bibitem[Zhao et~al.(2026)Zhao, Zhou, Li, Tang, Dong, Hou, Zhang, Min, Zhang, Liu, Wang, Du, Yang, Chen, Chen, Jiang, Ren, Li, Tang, Liu, Hu, Nie, and Wen]{Zhao_2026}
Wayne~Xin Zhao, Kun Zhou, Junyi Li, Tianyi Tang, Zican Dong, Yupeng Hou, Beichen Zhang, Yingqian Min, Junjie Zhang, Peiyu Liu, Xiaolei Wang, Yifan Du, Chen Yang, Yushuo Chen, Zhipeng Chen, Jinhao Jiang, Ruiyang Ren, Yifan Li, Xinyu Tang, Zikang Liu, Yiwen Hu, Jian-Yun Nie, and Ji-Rong Wen.
\newblock A survey of large language models.
\newblock \emph{Frontiers of Computer Science}, 20\penalty0 (12), May 2026.
\newblock ISSN 2095-2236.
\newblock \doi{10.1007/s11704-026-60308-3}.
\newblock URL \url{http://dx.doi.org/10.1007/s11704-026-60308-3}.

\end{thebibliography}
\bibliographystyle{iclr2027_conference}

\newpage
\appendix
\section{Appendix}

\subsection{More about Cross-tokenizer Distillation}
\label{app: related}
Compared with common OPD, the key challenge in cross-tokenizer distillation is that the teacher and student output distributions cannot be directly aligned \citep{song2026surveyonpolicydistillationlarge}.
Specifically, the two models exhibit misalignment in both their input tokenization and output logits.
Existing cross-tokenizer distillation approaches generally fall into two categories: (I) those that align the output distributions and (II) those that bypass this issue entirely by performing alignment in the representation space \citep{wang2025largelanguagemodelenhanced}.

For (I), we discuss existing approaches from the two aspects.
(1) For input-tokenization alignment. 
Some early studies use optimal transport or edit distances to measure pairwise distances between tokens \citep{le2025cot2aligncrosschainthoughtdistillation,cui2025multileveloptimaltransportuniversal,chen2025enhancingcrosstokenizerknowledgedistillation,DBLP:conf/aaai/VuongLTVL26}.
Later, one common way \citep{minixhofer2025universalcrosstokenizerdistillationapproximate,patiño2025_unlocking_on_policy_distillation_for_any_model_family,DBLP:conf/icml/Shin0LG25,nguyen2026ctpdcrosstokenizerpreference,sun2026simctrecoveringlostsupervision,sreenivas2026xtokenprojectionguidedcrosstokenizerknowledge,dao2026sraspanrepresentationalignment} is to identify shared text spans across different tokenizers and aggregate token-level distributions into span-level distributions using joint probability.
Another line of work converts tokens into byte sequences. 
It uses a prefix trie to marginalize token probabilities into next-byte conditional distributions, thereby aligning the teacher and student in a shared byte space \citep{phan2026crosstokenizerlikelihoodscoringalgorithms,singh2026crosstokenizerllmdistillationbytelevel,wang2026crosstokenizeronpolicydistillationbyteprefix}.
(2) For output-logit alignment, current strategies generally manipulate the output distributions through operations such as ranking \citep{cui2025multileveloptimaltransportuniversal,chen2025enhancingcrosstokenizerknowledgedistillation,DBLP:conf/aaai/WangJSWLRT26,DBLP:conf/aaai/VuongLTVL26,sreenivas2026xtokenprojectionguidedcrosstokenizerknowledge}, padding \citep{boizard2025crosstokenizerdistillationuniversallogit}, and key-token exploration \citep{DBLP:conf/icml/Shin0LG25,nguyen2026ctpdcrosstokenizerpreference} based on token probabilities, or divide the vocabularies into matched and unmatched tokens \citep{patiño2025_unlocking_on_policy_distillation_for_any_model_family,singh2026crosstokenizerllmdistillationbytelevel}, and identify a subset of shared logit dimensions across different vocabularies.

As for (II), current papers transfer knowledge in representation space through cross-space projection \citep{zhang2024dualspaceknowledgedistillationlarge}, sparse embedding reconstruction \citep{goddard2025trainingfreetokenizertransplantationorthogonal}, time-warped alignment \citep{vu2026dwakddualspaceweightingtimewarped}, or layer-wise hidden-state matching \citep{yang2026oprdonpolicyrepresentationdistillation}.

In this paper, we focus on category (I) and introduce the DVLM method to provide distribution-aligned supervision with the student’s input-tokenization and output-logit during OPD.

\subsection{Token Alignment Group}
\label{app: token alignment group}
We detail the construction procedure of Token Alignment Groups, with the pseudocode provided in Algorithm~\ref{alg:token_alignment}.
The algorithm first converts the student and teacher token sequences into canonical text pieces by incrementally decoding each token prefix and extracting the newly added substring. 
It then maintains a text buffer for each tokenizer and greedily extends the shorter buffer until the two decoded strings match. Once a match is found, the corresponding student and teacher token indices are recorded as an alignment group, and both buffers are reset. 
Finally, any remaining token indices are retained as a residual group to ensure complete sequence coverage.

Taking the input sentence in Figure~\ref{fig:main} as an example, 
when $k = 1$, $\mathcal{G}_1^T = \left[`distillation'\right]$ and $\mathcal{G}_1^S = \left[`distill', `ation'\right]$, the two share the same text span `distillation';
when $k = 2$, $\mathcal{G}_2^T = \left[`is'\right]$ and $\mathcal{G}_2^S = \left[`is'\right]$, the two share the same text span `is';
when $k = 3$, $\mathcal{G}_3^T = \left[`challeng', `ing.'\right]$ and $\mathcal{G}_3^S = \left[`challenging.'\right]$, the two share the same text span `challenging.'.

\begin{algorithm}[t]
\caption{Greedy Construction of Token Alignment Groups}
\label{alg:token_alignment}
\begin{algorithmic}[1]

\Require Student tokens $\mathbf{s}$, teacher tokens $\mathbf{t}$, and tokenizers $\tau_S,\tau_T$
\Ensure Alignment groups $\mathcal{A}=\{(S_k,T_k)\}_{k=1}^{K}$

\Function{IncrementalDecode}{$\tau,\mathbf{z}$}
    \State $\mathbf{p}\gets[\,]$; $v\gets\varepsilon$
    \For{$k=1$ to $\Call{Length}{\mathbf{z}}$}
        \State $u\gets\Call{Decode}{\tau,\mathbf{z}_{1:k}}$
        \State $p_k\gets u[\Call{Length}{v}+1:]$
        \Comment{newly decoded suffix}
        \State $\mathbf{p}\gets\Call{Concat}{\mathbf{p},[p_k]}$; $v\gets u$
    \EndFor
    \State \Return $\mathbf{p}$
\EndFunction

\State $\mathbf{p}^S\gets\Call{IncrementalDecode}{\tau_S,\mathbf{s}}$
\State $\mathbf{p}^T\gets\Call{IncrementalDecode}{\tau_T,\mathbf{t}}$
\State $i,j\gets1$; $B_S,B_T\gets\varepsilon$; $S,T,\mathcal{A}\gets[\,]$

\While{$i\le L_S$ \textbf{or} $j\le L_T$}
    \If{$B_S=B_T$ \textbf{and} $B_S\neq\varepsilon$}
        \State $\mathcal{A}\gets\Call{Concat}{\mathcal{A},[(S,T)]}$
        \State $B_S,B_T\gets\varepsilon$; $S,T\gets[\,]$
        \State \textbf{continue}
    \EndIf

    \If{$B_S=\varepsilon$ \textbf{and} $i\le L_S$}
        \State $\sigma\gets S$
    \ElsIf{$B_T=\varepsilon$ \textbf{and} $j\le L_T$}
        \State $\sigma\gets T$
    \ElsIf{$(\Call{Length}{B_S}\le\Call{Length}{B_T}\ \textbf{and}\ i\le L_S)\ \textbf{or}\ j>L_T$}
        \State $\sigma\gets S$
    \Else
        \State $\sigma\gets T$
    \EndIf

    \If{$\sigma=S$}
        \State $B_S\gets\Call{Concat}{B_S,p_i^S}$; $S\gets\Call{Concat}{S,[i]}$; $i\gets i+1$
    \Else
        \State $B_T\gets\Call{Concat}{B_T,p_j^T}$; $T\gets\Call{Concat}{T,[j]}$; $j\gets j+1$
    \EndIf
\EndWhile

\If{$S\neq[\,]$ \textbf{or} $T\neq[\,]$}
    \State $\mathcal{A}\gets\Call{Concat}{\mathcal{A},[(S,T)]}$
\EndIf

\State \Return $\mathcal{A}$

\end{algorithmic}
\end{algorithm}

\subsection{More about our Methods}

In the PTS, position IDs represent logical text positions rather than physical indices in the concatenated sequence. 
Native teacher tokens retain their original positions. 
Each re-tokenized token is positioned based on the teacher prefix preceding its alignment group, the preceding expanded tokens within that group, and its offset in the current expansion. 
Thus, despite physical concatenation, the tokens preserve their aligned logical positions under the HPA attention mask.

Meanwhile, since Llama-3 and OLMo-2 use byte-level BPE tokenizers, a single token may represent only part of a multibyte character. 
We therefore adopt a data-level filtering strategy: during preprocessing, we discard any sample containing a student token that cannot be independently decoded and correctly re-tokenized. 
This ensures that every student token retained for PTS construction can be safely converted into a valid sequence of teacher tokens.

\subsection{Supplementary Training Settings}
\paragraph{Models.}
In this work, we use Qwen3-4B as the teacher model and Llama-3.2-1B-Instruct and OLMo-2-1B-Instruct as the student models.
The vocabulary sizes of the three models are $151,669$, $128,256$, and $100,278$, respectively. 
Therefore, OLMo-2-1B-Instruct exhibits a larger vocabulary gap with the teacher model, Qwen3-4B, which theoretically makes distillation more challenging.

\paragraph{Data.}
For the training of the new LM heads, we randomly sample $1$B tokens from a single parquet file in the High-Quality-Synthetic subset of Nemotron-CC-v2, and randomly select $0.05$B tokens of mathematical data and $0.05$B tokens of code data from Skywork-OR1-Data based on their category labels. 
These data are then mixed to form a total training corpus of $1.1$B tokens.
In practice, the models are able to converge after processing only about one-fifth of these tokens, corresponding to approximately $0.22$B training tokens.
The primary purpose of mixing these datasets is to expose the new LM heads to as many diverse tokens as possible.
For DVLM-based OPD, we also mix the DAPO-Math (math-related) and the TACO (code-related) to construct the training set.
For DAPO-Math, we follow a publicly available HuggingFace dataset page\footnote{https://huggingface.co/datasets/mnoukhov/dapo-math-17k-processed-filtered-qwen3-4b-base-32samples/viewer/default/train} in which each problem is rolled out $32$ times using our teacher Qwen3-4B and the corresponding pass rate is recorded. 
We select all problems with a pass rate of at least $25$\%, around $3,863$ samples.
For TACO, we select problems labeled as `EASY' or `MEDIUM', around $3,954$ samples. 
In total, the number of our training samples is $7,817$.
The experiments in Section~\ref{sec: diff} demonstrate that, although our training set is relatively small and consists of moderately challenging examples, it is well suited to the two student models considered in this work.

\paragraph{Training Implementation.}
All experiments are conducted on $4$ NVIDIA H$200$ GPUs, each with $141$ GB of memory. 
For Stage I, we use only one GPU. 
For Stage II, we deploy the student model with vLLM \citep{kwon2023efficient} on one GPU to generate responses for OPD and refresh its weights as the student parameters are updated. 
The remaining three GPUs are used for data-parallel training.
The code for II is adapted from HuggingFace’s TRL library.
The training hyperparameters for these two stages are provided in Table~\ref{tab:training_para}.

\begin{table}[t]
\centering
\captionsetup{skip=1.5pt}
\renewcommand{\arraystretch}{1.2}
\small
\caption{Training configurations for the new LM head and the GOLD baseline.}
\label{tab:training_para}
\begin{tabular}{lcc}
\toprule
 & Stage I & Stage II \\
\midrule
Training Objective & next-token prediction & forward KL \\
Learning Rate & $1\times10^{-5}$ & $2\times10^{-6}$ \\
Batch Size & $48$ & $64$ \\
Gradient Accumulation steps & $1$ & $1$ \\
Training Epochs & $1$ & $2$ \\
Learning-rate Schedule & Cosine & Cosine \\
Warmup Ratio & $0.10$ & $0.01$ \\
Weight Decay & $0.01$ & -- \\
Gradient Clipping & $1.0$ & -- \\
Numerical Precision & BF16 teacher, FP32 head & BF16 \\
Maximum Sequence length & $4096$ & $2048$ \\
Maximum Completion length & -- & $1024$ \\
Logging Interval (steps) & $10$ & $10$ \\
Checkpoint Interval (steps) & $1000$ & $10$ \\
\bottomrule
\end{tabular}
\end{table}

\begin{table}[t]
\centering
\captionsetup{skip=1.5pt}
\renewcommand{\arraystretch}{1.2}
\small
\caption{Evaluation settings for the four understanding tasks.}
\label{tab:understanding_eval_settings}
\begin{tabular}{ll}
\toprule
\textbf{Settings} &  \\
\midrule
Model Interface & new LM head \\
Teacher Model & Qwen3-4B \\
Prediction Format & single-token generation \\
Shot & $0$ \\
Batch Size & $32$ \\
Numerical Precision & BF16 \\
\bottomrule
\end{tabular}
\end{table}

\begin{table*}[t]
\centering
\captionsetup{skip=1.5pt}
\small
\renewcommand{\arraystretch}{1.2}
\caption{Evaluation settings for the LM-Evaluation-Harness framework.}
\label{tab:LM-Evaluation-Harness}
\begin{tabularx}{\textwidth}{lcccccc}
\toprule
\textbf{Task} & \textbf{Batch Size} & \textbf{Shots} & \textbf{Max.\ Gen.\ Tokens} & \textbf{Precision} & \textbf{Temp.} & \textbf{Metrics} \\
\midrule
GSM8K & $32$ & $0$ & $1024$ &  BF16 & $0.0$ & ACC. \\
MATH-500 & $16$ & $0$ & $1024$ & BF16 & $0.0$ & ACC.\\
HumanEval+ & $32$ & $0$ &  $1024$ & BF16 & $0.0$ & pass@$1$(n=$1$)\\
CRUXEval & $32$ &$2$ &  $1024$ & BF16 & $0.2$ & pass@$1$(n=$1$)\\
\bottomrule
\end{tabularx}
\end{table*}

\begin{table*}[t]
\centering
\captionsetup{skip=1.5pt}
\small
\renewcommand{\arraystretch}{1.2}
\caption{Evaluation settings for the Evalchemy framework.}
\label{tab:Evalchemy}
\begin{tabularx}{\textwidth}{lcccccc}
\toprule
\textbf{Task} & \textbf{Batch Size} & \textbf{Shots} & \textbf{Max.\ Gen.\ Tokens} & \textbf{Precision} & \textbf{Temp.} & \textbf{Metrics} \\
\midrule
AMC23 & $16$ & $0$ & $1024$ &  BF16 & $0.0$ & pass@$1$(n=$16$) \\
LiveCodeBench v5& $16$ & $0$ & $1024$ & BF16 & $0.0$ & pass@$1$(n=$8$)\\
\bottomrule
\end{tabularx}
\end{table*}

\subsection{Supplementary Evaluation Settings}
\label{app: evaluation settings}

\paragraph{Benchmarks.} 
Most of the content is presented in the main text.
For MMLU, we choose the subset MMLU (abstract algebra), named `MMLU (Algebra)'.

\paragraph{Baselines.} We consider two categories of baselines: common LLM reasoning enhancement methods, including SFT and RL, specifically GRPO; and four types of cross-tokenizer distillation methods: DSKD (representation-based), ALM (span-based), GOLD (from which our code is adapted), and BPM (byte-based).
Our code is adapted from the GOLD implementation in HuggingFace’s TRL library. 
We add DVLM training and configuration. 
During cross-tokenizer distillation, we construct PTS inputs for the teacher, apply HPA, and extract teacher logits at the corresponding positions.
The code for all four cross-tokenizer distillation frameworks is publicly available. 
To ensure a fair comparison, we reproduce their training results using our own training settings.

\paragraph{Metrics.}
AMC23 and LiveCodeBench v5 are evaluated using Evalchemy\footnote{https://github.com/mlfoundations/evalchemy}, while all other benchmarks in Stage I and Stage II are tested with LM-Evaluation-Harness\footnote{https://github.com/eleutherai/lm-evaluation-harness}.
For the four understanding tasks, as well as GSM8K and MATH-500, we utilize accuracy as the evaluation metric; for AMC23, HumanEval+, CRUXEval, and LiveCodeBench v5, we apply pass@$1$.

All evaluations are conducted on a single H200 GPU.
For the four understanding tasks, we modify their evaluation settings in the LM-Evaluation-Harness framework to use single-token prediction.
We report the evaluation settings for the four understanding tasks, the LM-Evaluation-Harness framework, and the Evalchemy framework in Table~\ref{tab:understanding_eval_settings}, Table~\ref{tab:LM-Evaluation-Harness}, and Table~\ref{tab:Evalchemy}, respectively.

\begin{table*}[t]
\centering
\captionsetup{skip=1.5pt}
\caption{Trainable parameters and training time of our method.}
\label{tab:training_cost}
\resizebox{\linewidth}{!}{
\begin{tabular}{llccc}
\toprule
\textbf{Stage}
& \textbf{Components}
& \textbf{Runs}
& \textbf{Trainable Parameters}
& \textbf{Days} \\
\midrule

\multirow{2}{*}{Stage I}
& Llama head DVLM & $1 \times 3$ & $0.33$B & $0.08$ (single GPU)\\
& OLMo head DVLM & $1 \times 3$ & $0.26$B & $0.09$ (single GPU)\\
\midrule
\multirow{2}{*}{Stage II}
& Llama head DVLM $\rightarrow$ Llama & $1 \times 3$ & $1.24$B & $0.23$ (four GPUs)\\
& OLMo head DVLM $\rightarrow$ OLMo & $1 \times 3$ & $1.48$B & $0.27$ (four GPUs)\\

\bottomrule
\end{tabular}
}
\end{table*}

\subsection{Supplementary Experiments}
\label{app: supplementary experiments}

\subsubsection{Trainable Parameters, Training Runs, and Training Variance}
\label{app: more training}
We report the number of trainable parameters, training runs, and training time for each stage of DVLM in Table~\ref{tab:training_cost}. 
Training time is measured in (wall-clock) Days.
We run experiments with three random seeds in parallel on a single GPU in Stage I.
For Stage I, we use the step at which training converges as the cutoff.
As shown in the table, the two LM heads in Stage I increase the number of trainable parameters by 26\% and 18\%, respectively, while introducing only a small amount of additional training time.
Accounting for the same number of GPUs used, Stage I adds only about 8\% to the same-GPU hour cost for both teacher–student pairs.
Therefore, Stage I can be considered lightweight.
To reduce variance caused by training randomness, we run Stage II three times with different random seeds and report the averaged results in Table~\ref{tab:main_results}.

Accordingly, we report the variation in the main experimental results. 
Notably, we report variance only for baselines that can be trained.
Overall, for a given student model and benchmark, the variances are similar across different methods.
For each reported value, we take the larger absolute value of the upper and lower variance bounds and round it to one decimal place.

\begin{table}[t]
\centering
\captionsetup{skip=1.5pt}
\small
\caption{Held-out language-modeling evaluation on unseen AI-generated text.}
\label{tab:teacher_bpb}
\begin{tabular}{lcc}
\toprule
Teacher head
& mean token loss
& BPB $\downarrow$ \\
\midrule
Original Qwen head & $1.94$  & $\approx 0.37$ \\
DVLM (Llama vocabulary) & $1.93$   & $\approx 0.37$ \\
DVLM (OLMo vocabulary)  & $1.90$   & $\approx 0.36$ \\
\bottomrule
\end{tabular}
\end{table}

\begin{table}[t]
\centering
\captionsetup{skip=1.5pt}
\small
\caption{The evaluation of the DVLM teachers on Math-500 and Humaneval+.}
\label{tab:teacher_reasoning}
\begin{tabular}{lcc}
\toprule
Teacher head
& Math-500
& Humaneval+ \\
\midrule
Original Qwen head & $68.40$  & $82.93$ \\
DVLM (Llama vocabulary) & $66.80$   & $82.32$ \\
DVLM (OLMo vocabulary)  & $68.60$   & $81.71$ \\
\bottomrule
\end{tabular}
\end{table}

\subsubsection{More about Ablation Study}
We provide a detailed description of the ablation settings.
(1) `w/o HPA’: Retrain the LM head, apply standard causal attention to the concatenated input, and extract the student-token logits at the corresponding positions. 
(2) `w/o T.Prefix’: Retrain the LM head using only the student re-tokenized groups as input, and extract the student-token logits at the corresponding positions.
(3) `w/o Re-token.’: Retrain the LM head using only the teacher-tokenized sequence as input, without logit extraction. 
For student tokens absent from the teacher vocabulary, assign their weights by averaging the trained LM-head weights of their re-tokenized teacher tokens in the end.
(4) `w/o LM Head’: Use the original teacher model and follow GOLD in Stage II to handle matched and unmatched logits separately.
(5) `w/o L.Extract: In Stage II, unaligned token groups are handled using GOLD’s span-level method.

We also find that randomly initializing the new LM head prevents it from converging to the original teacher model’s loss level.

\subsubsection{The Extra Evaluation of DVLM Architecture}
\label{app: dvlm per}
\paragraph{The bits per byte of the new LM heads}
To compare the original teacher with the two adapted DVLM teachers across different tokenizations, we sample $500$ held-out passages with length around $4$K generated by Qwen3.8-27B that are excluded from LM-head training. 
We evaluate all three teachers on the same text using teacher-forced next-token prediction: the original teacher used its native tokenizer and LM head, while each DVLM used its corresponding student vocabulary. 
We report both mean token-level loss and bits per byte (BPB) \citep{liu2025superbpespacetravellanguage}, where BPB is the total NLL divided by the number of UTF-8 bytes in the scored text and by ($\ln 2$). 
Lower BPB indicates better predictive performance on the same text after accounting for differences in tokenization. 
As measured by BPB, Table~\ref{tab:teacher_bpb} shows that both DVLM heads achieve language-modeling performance comparable to the original teacher.

\paragraph{The reasoning performances of the new LM heads}
We also evaluate the two new DVLM teachers on two reasoning benchmarks, MATH-500 and HumanEval+. 
Because the DVLM teacher employs different input and output tokenizers, existing generation and evaluation frameworks are not directly applicable. 
We therefore implement a custom autoregressive generation pipeline using HuggingFace. 
At each decoding step, the model generates one token, which is decoded into text, appended to the existing sequence, and then re-tokenized as input for the next step. 
Finally, all generated responses are evaluated using LM-Evaluation-Harness. 
The results in Table~\ref{tab:teacher_reasoning} show that the DVLM teachers achieve performance comparable to the original teacher, exhibiting only a minor performance drop.

\subsubsection{Training Time Comparison with GOLD}
Since our implementation is adapted from GOLD algorithm in the TRL library, we compare the training time of DVLM with that of GOLD. 
For the Llama-3.2-1B and OLMo-2-1B backbones, GOLD requires $0.35$ and $0.45$ days, respectively, substantially longer than DVLM. 
This difference mainly arises because GOLD must process the logits of matched and unmatched tokens separately, whereas DVLM directly computes the loss over teacher and student logits with identical dimensions.

\subsubsection{Comparison with Same-tokenizer Distillation}
For the OLMo-2-1B-Instruct student, we compare Qwen3-4B, the cross-tokenizer teacher applied in this paper, with an additional same-tokenizer teacher, OLMo-2-7B-Instruct.
For same-tokenizer distillation, we directly use forward KL divergence as the optimization objective.
The results in Table~\ref{tab:same_token} show that our method outperforms same-tokenizer distillation on five of the six benchmarks, with LiveCodeBench v5 being the exception. 
This is likely because our cross-tokenizer teacher, Qwen3-4B, is substantially stronger than OLMo-2-7B-Instruct. 
A practical implication is that, when a sufficiently capable teacher is available, cross-tokenizer OPD could be a viable choice.

\begin{table}[t]
\centering
\captionsetup{skip=1.5pt}
\caption{Comparison of same-tokenizer and cross-tokenizer distillation for OLMo-2-1B-Instruct.}
\fontsize{9pt}{9.5pt}\selectfont
\setlength{\tabcolsep}{3.0pt}
\renewcommand{\arraystretch}{1.2}
\begin{tabularx}{\linewidth}{
>{\raggedright\arraybackslash}p{1.8cm}
*{7}{>{\centering\arraybackslash}X}
}
\toprule
\multirow{3}{*}{{\fontsize{10pt}{12pt}\selectfont Method}} & \multicolumn{3}{c}{{\fontsize{10pt}{12pt}\selectfont Mathematics}} & \multicolumn{3}{c}{{\fontsize{10pt}{12pt}\selectfont Coding}} & \multirow{3}{*}{{\fontsize{10pt}{12pt}\selectfont Average}} \\
\cmidrule(lr){2-4} \cmidrule(lr){5-7}
& GSM8K & MATH-500 & AMC23 & HumanEval+ & CRUXEval & \makecell{LCBench} & \\
& ACC. & ACC. & pass@1 & pass@1 & pass@1 & pass@1 &  \\
\midrule
\rowcolor{HeaderOrange} OLMo-2-1B & $69.37$ & $21.00$ & $5.00$ & $23.17$ & $12.31$ & $0.27$ & $21.85$ \\

\midrule
\rowcolor{TeacherBlue} Qwen3-4B & $87.79$ & $68.40$ & $38.28$ & $82.93$ & $41.81$ & $26.93$ & $57.69$ \\
\textbf{DVLM} & \best{$70.15$} & \best{$22.87$} & \best{$8.39$} & \best{$27.44$} & \best{$17.94$} & $1.50$ & \best{$24.72$} \\
\midrule
\rowcolor{TeacherBlue} OLMo-2-7B & $80.36$ & $35.40$ & $7.57$ & $25.60$ & $16.06$ & $6.90$ &  $28.65$\\
\textbf{Forward KL} & $69.97$ & $21.80$ & $6.58$ & $24.39$ & $14.68$ & \best{$1.85$} & $23.21$ \\
\bottomrule
\end{tabularx}
\label{tab:same_token}
\end{table}

\subsubsection{Comparison with Tokenizer Transfer Method}
To determine whether our method can be addressed by conventional tokenizer transfer methods, we select TokAlign \citep{li-etal-2025-tokalign}, which substantially modifies the teacher model by not only retraining its LM head but also completely replacing its vocabulary, token embeddings, and input tokenization scheme. 
We reproduce its results under our experimental setting on Llama-3.2-1B-Instruct.
As shown in Table~\ref{tab:tokalign_comparison}, DVLM consistently outperforms TokAlign across all four benchmarks.
The results show that the conventional tokenizer-transfer method TokAlign substantially degrades student performance in cross-tokenizer distillation on challenging reasoning tasks. 
This can be explained from two perspectives. 
First, the original TokAlign paper primarily evaluates language-understanding tasks, leaving its effectiveness on reasoning-intensive tasks unclear. 
Second, TokAlign replaces the teacher’s input tokenization scheme, creating a substantial pretraining–adaptation mismatch and preventing its pretrained parameters from being fully utilized. 
In contrast, DVLM trains only an additional LM head while preserving the teacher’s native input interface and backbone, thereby retaining its underlying capabilities to the greatest possible extent.

\begin{table}[t]
\centering
\captionsetup{skip=1.5pt}
\caption{Comparison of the TokAlign transfer method and our DVLM method.}
\label{tab:tokalign_comparison}
\fontsize{9pt}{9.5pt}\selectfont
\setlength{\tabcolsep}{3.0pt}
\renewcommand{\arraystretch}{1.2}
\begin{tabularx}{\linewidth}{
>{\raggedright\arraybackslash}p{1.8cm}
*{5}{>{\centering\arraybackslash}X}
}
\toprule
\multirow{3}{*}{{\fontsize{10pt}{12pt}\selectfont Method}}
& \multicolumn{2}{c}{{\fontsize{10pt}{12pt}\selectfont Mathematics}}
& \multicolumn{2}{c}{{\fontsize{10pt}{12pt}\selectfont Coding}}
& \multirow{3}{*}{{\fontsize{10pt}{12pt}\selectfont Average}} \\
\cmidrule(lr){2-3} \cmidrule(lr){4-5}
& MATH-500 & AMC23 & HumanEval+ & \makecell{LCBench} & \\
& ACC. & pass@1 & pass@1 & pass@1 & \\
\midrule

\rowcolor{TeacherBlue}
Qwen3-4B
& $68.40$ & $38.28$ & $82.93$ & $26.93$ & $54.14$ \\

\midrule

\rowcolor{HeaderOrange}
Llama-3.2-1B
& $27.00$ & $4.68$ & $31.10$ & $1.47$ & $16.06$ \\

TokAlign
& $24.40$
& $2.82$
& $18.29$
& $0.57$
& $11.52$ \\

\textbf{DVLM}
& \best{$29.93$}
& \best{$9.06$}
& \best{$36.99$}
& \best{$2.86$}
& \best{$19.71$} \\

\bottomrule
\end{tabularx}
\end{table}

\subsection{Limitations}
Due to limited GPU resources and the need to consider the baseline performance of both teacher and student models, we could not conduct more extensive experiments with larger models.

\end{document}